%% file: neurips_2026_camera.tex
\documentclass{article}
\PassOptionsToPackage{numbers}{natbib}
\usepackage[eandd, final]{neurips_2026}
\usepackage[utf8]{inputenc}
\usepackage[T1]{fontenc}
\usepackage{hyperref}
\usepackage{url}
\usepackage{booktabs}
\usepackage{amsmath,amssymb,amsfonts}
\usepackage{graphicx}
\usepackage{textcomp}
\usepackage{microtype}
\usepackage{xcolor}
\usepackage{array}
\usepackage{multirow}
\usepackage{longtable}
\usepackage{placeins}

\title{FlyingDrones: A Dataset and Benchmark for Optical Flow Estimation from UAV motion}

\author{%
Laurenz Ruzicka \\
Digital Safety and Security \\ 
Data Science and Artificial Intelligence \\
Austrian Institute of Technology \\ 
\texttt{Laurenz.Ruzicka@ait.ac.at} \\
\And
Roman Pflugfelder \\
Digital Safety and Security \\ 
Data Science and Artificial Intelligence \\
Austrian Institute of Technology \\ 
School of Computation, Information and Technology \\
Technical University of Munich \\
Munich Center for Machine Learning (MCML) \\
\texttt{Roman.Pflugfelder@\{ait.ac.at|tum.de\}} \\
\And
Daniel Cremers \\
School of Computation, Information and Technology \\
Technical University of Munich \\
Munich Center for Machine Learning (MCML)\\
\texttt{cremers@tum.de} \\
}

\begin{document}

\maketitle

\begin{abstract}
Optical flow is an important cue for drone analysis in aerial surveillance, airspace monitoring, and long-range target tracking, where motion remains informative when appearance is weak. Estimating the motion of small Unmanned Aerial Vehicles (UAVs), however, is difficult because distant targets occupy few pixels, have low contrast, and move against complex dynamic backgrounds. Standard optical-flow benchmarks emphasize large, object-centric motions and therefore do not characterize distant aerial targets well. We introduce FlyingDrones, a generated benchmark with dense ground-truth optical flow and segmentation masks derived from physically based rendering, and define resolution-normalized metrics (EPE Norm) that normalize flow by image dimensions. We evaluate 16 state-of-the-art model families (88 public checkpoints) on a motion- and size-stratified subset of 999 flow pairs at three resolutions, with bootstrap confidence intervals and a zero-flow reference. Every family recovers drone motion better than the zero-flow reference, but the gap between families widens with resolution: at $1920\times1080$, global-matching, recurrent, transformer, and short-window temporal models separate clearly from pyramid, classical, self-supervised, lightweight, and diffusion models. Within-family checkpoint spread is comparable to or larger than between-family spread, so the training curriculum is as important as the architecture. Two complementary studies test the benchmark beyond synthetic accuracy: a ground-truth-free motion-alignment evaluation on 21 cameras of real UAV footage shows that the synthetic ranking carries a moderate, significant signal about real-world performance (Spearman $\rho=0.63$ over 41 checkpoints) while the exact top ranks differ, and a fine-tuning study shows that 800 FlyingDrones pairs reduce held-out drone-region error by up to $37\%$ and improve drone localization on the real footage for all four architectures tested.
\end{abstract}

\section{Introduction}

Apparent motion is an important visual cue in drone analysis and aerial surveillance~\cite{srigrarom2020, mahalanobis2025}. At long ranges, drones often occupy only a few pixels and lack stable features due to simple appearance textures and complex backgrounds. This means that the relevant signal is weak, localized, and additionally often superimposed by camera-induced background motion, making motion computation challenging.

This work therefore evaluates and compares a broad range of optical flow methods under the tiny motion regime: classical and early deep baselines such as PyFlow~\cite{pathak_learning_2017}, FlowNet2~\cite{ilg2017}, PWC-Net~\cite{sun2018}, LiteFlowNet3~\cite{hui_liteflownet3_2020}, and VCN~\cite{yang_volumetric_2019}; recurrent correlation-based models such as RAFT~\cite{teed2020} and SEA-RAFT~\cite{wang2025a}; lightweight or self-/unsupervised variants such as FastFlowNet~\cite{kong_fastflownet_2021} and ARFlow~\cite{liu_learning_2020}; transformer-style correspondence models such as GMFlow~\cite{xu_gmflow_2022}, UniMatch~\cite{xu_unifying_2023}, FlowFormer++~\cite{shi_flowformer_2023}, and WAFT~\cite{wang2025}; temporally extended models such as VideoFlow~\cite{shi_videoflow_2023} and MemFlow~\cite{dong_memflow_2024}; and the generative FlowDiffusion~\cite{saxena_surprising_2023, dong_open-ddvm_2023}. 

State of the art (SOTA) datasets and metrics account to the uncertainty in the behavior of optical flow methods with tiny motions: FlyingChairs~\cite{dosovitskiy2015} and FlyingThings3D~\cite{mayer_large_2016} provide large-scale synthetic supervision, Sintel~\cite{Butler:ECCV:2012} emphasizes cinematic effects and non-rigid motion, KITTI~\cite{Menze2015CVPR} focuses on outdoor driving scenes, and TartanAir~\cite{tartanair2020iros} broadens camera trajectories and environments. None of the existing datasets target drones, tiny aerial objects or in general tiny motion regions. Standard measures such as end-point error (EPE), 1px, 3px, 5px, and Fl-all are informative for moderate or large motion regions but become hard to interpret when the target occupies only a few pixels.

To address these issues, we introduce FlyingDrones, a synthetically generated benchmark for optical flow assessment of drone motion, with dense optical-flow and segmentation ground truth across controlled image resolution regimes. The paper also proposes drone-pixel-only metrics and a new resolution-normalized EPE, which makes flow components independent of image width and height, and use them to study SOTA optical flow methods and training curricula independently of image resolution. To the best of our knowledge, this work is the first systematic assessment of optical flow methods applied to tiny motion regions like those created by drones.

To summarize, this papers makes the following contributions:
\begin{itemize}
    \item We introduce FlyingDrones, a synthetically generated dataset with dense optical-flow and segmentation ground truth for evaluating drone motion across target sizes and image resolutions.
    \item We define resolution-normalized evaluation measures, especially EPE Norm, to make cross-resolution performance more interpretable.
    \item We compare 16 optical-flow methods (88 public checkpoints) on a motion- and size-stratified subset of the benchmark with confidence intervals and a zero-flow reference, and show that the families separate increasingly with resolution, that within-family checkpoint spread rivals between-family spread, and that Sintel-style fine-tuning increasingly outperforms KITTI-style fine-tuning as resolution grows.
    \item We validate the benchmark beyond synthetic accuracy: a ground-truth-free motion-alignment study on 21 cameras of real UAV footage~\cite{li_reconstruction_2020} measures how far the synthetic ranking transfers to real data, and a fine-tuning study on a held-out FlyingDrones split shows that the dataset improves both held-out synthetic accuracy and real-footage drone localization.
\end{itemize}

\section{Related Work}

Classical optical flow methods such as Horn--Schunck~\cite{horn1981} and Lucas--Kanade~\cite{lucas1981} formulate motion estimation as a variational problem with brightness-constancy and smoothness assumptions, but struggle with large displacements, strong appearance changes, and complex non-rigid motion.
Early deep optical-flow models replaced handcrafted optimization with supervised end-to-end learning. FlowNet and FlowNet2~\cite{dosovitskiy2015, ilg2017} established the Chairs/Things pre-training pipeline used by many later methods, while coarse-to-fine architectures such as PWC-Net~\cite{sun2018}, LiteFlowNet/LiteFlowNet3~\cite{hui18liteflownet, hui_liteflownet3_2020}, and FastFlowNet~\cite{kong_fastflownet_2021} combined feature pyramids, warping, and compact cost volumes for efficient inference. These models perform well on conventional benchmarks, but repeated downsampling can suppress very small motion cues.
A related family emphasizes richer matching representations. VCN~\cite{yang_volumetric_2019} uses volumetric correspondence reasoning; RAFT~\cite{teed2020} replaces coarse-to-fine refinement with a dense all-pairs correlation volume and iterative updates at fixed feature resolution; SEA-RAFT~\cite{wang2025a} simplifies RAFT and models uncertainty more explicitly; and WAFT~\cite{wang2025} uses warping-based alignment with a transformer-style update module. These methods are especially relevant here because they preserve finer spatial information than traditional pyramid approaches.
ARFlow~\cite{liu_learning_2020} learns optical flow with self-supervised consistency constraints, and PyFlow remains a useful non-deep variational baseline. Such methods are attractive when labels are scarce or speed matters, but often trade accuracy for robustness or efficiency.
GMFlow~\cite{xu_gmflow_2022}, UniMatch~\cite{xu_unifying_2023}, and FlowFormer++~\cite{shi_flowformer_2023} model long-range interactions more explicitly than local convolutional pipelines. VideoFlow~\cite{shi_videoflow_2023} uses multi-frame inputs, MemFlow~\cite{dong_memflow_2024} adds an explicit memory mechanism, and FlowDiffusion~\cite{saxena_surprising_2023, dong_open-ddvm_2023} formulates prediction with a diffusion-style process.
Across these categories, models differ in supervision, matching representation, temporal context, and computational budget, but most were developed on datasets dominated by moderate or large motions. Our benchmark studies their behavior when the relevant signal is the weak motion of a tiny aerial target.

Small aerial-target analysis is usually framed as object detection. Appearance-based detectors such as YOLO~\cite{redmon2016a} and Faster R-CNN~\cite{ren2017} are widely used, but their performance degrades when targets shrink to a few pixels because discriminative features are attenuated or lost during pooling and strided convolutions~\cite{lin_feature_2017, tong_recent_2020, liu_survey_2021, cheng_towards_2023}. Motion estimation can therefore provide complementary information~\cite{sun_enhancing_2023, alkentar_utilize_2024}.
Early motion-based strategies relied on background subtraction, which assumes a static viewpoint and fails under the ego-motion typical of aerial surveillance. Deep optical flow instead models dense pixel-wise displacements and can separate object motion from camera-induced parallax, but objects occupying only a fraction of a percent of the image generate weak, localized signals that are easily overwhelmed by background flow or smoothed out by spatial regularization and coarse-to-fine downsampling~\cite{sun2018}.
Despite the importance of optical flow for small-target analysis, empirical validation remains limited by the lack of suitable data.
\section{Metrics}\label{sec:methodology}

Let an image pair $(I_t, I_{t+1})$ of spatial resolution $W \times H$ be given, where $W$ and $H$ denote the image width and height in pixels. The optical flow field is defined as a dense vector field

\begin{equation}
    \mathbf{F} : \Omega \rightarrow \mathbb{R}^2, \quad \Omega = \{1,\dots,W\} \times \{1,\dots,H\},
\end{equation}

where $\Omega = \{1,\dots,W\} \times \{1,\dots,H\}$ denotes the discrete image domain and $\mathbf{F}(x,y) = (u(x,y), v(x,y))$ represents the horizontal and vertical displacement of pixel $(x,y)$ between frames. Let $\mathbf{F}_{gt}$ and $\mathbf{F}_{pred}$ denote the ground-truth and predicted flow fields, respectively. All metrics are computed per pixel and then averaged over the evaluation region.

\subsection{End-Point Error (EPE)}
The most commonly used metric in optical flow evaluation is the End-Point Error (EPE), defined at each pixel as the Euclidean distance between predicted and ground-truth flow vectors, 

\begin{equation}
    \mathrm{EPE}(x,y) = \left\| \mathbf{F}_{pred}(x,y) - \mathbf{F}_{gt}(x,y) \right\|_2,
\end{equation}
where $\|\cdot\|_2$ denotes the Euclidean ($\ell_2$) norm in $\mathbb{R}^2$. The reported EPE is the mean of this quantity over all evaluated pixels. 

\subsection{Pixel Threshold Metrics}
To complement EPE, we report pixel-threshold exceedance metrics denoted by $\mathrm{Px}_\tau$. These metrics compute the percentage of pixels whose endpoint error exceeds a fixed threshold $\tau$,
\begin{equation}
    \mathrm{Px}_\tau = \frac{1}{|\Omega|} \sum_{(x,y)\in\Omega} \mathbb{I}\left[ \mathrm{EPE}(x,y) > \tau \right],
\end{equation}
with $\tau \in \{1, 3, 5\}$ pixels and $\mathbb{I}[\cdot]$ the indicator function.
$\mathrm{Px}_1$ captures how often the error exceeds one pixel, while $\mathrm{Px}_3$ and $\mathrm{Px}_5$ report progressively coarser failure rates. These thresholds implicitly assume that fixed pixel errors are equally meaningful across resolutions and object sizes, which is not true for tiny drones.
This dependence is stronger in our benchmark because the same scene is rendered at different resolutions. At 480$\times$270, a 1px error corresponds to about $0.21\%$ of the image width and $0.37\%$ of the image height; at 1920$\times$1080, it is only about $0.05\%$ and $0.09\%$.

\subsection{Flow Outliers (FL)}
Flow Outliers (FL) are designed to account for relative motion magnitude by labeling a pixel as an outlier if its error is both absolutely large and large relative to the ground-truth displacement. A pixel is classified as an outlier if
\begin{equation}
    \mathrm{EPE}(x,y) > 3.0
    \quad \text{and} \quad
    \frac{\mathrm{EPE}(x,y)}{\left\| \mathbf{F}_{gt}(x,y) \right\|_2} > 0.05.
\end{equation}

The FL score is the percentage of pixels satisfying this condition.

\subsection{Resolution-Normalized and Drone Region Metrics}
To reduce the dependence of flow evaluation on image resolution and object scale, we introduce resolution-normalized variants of the standard metrics. Let $S = (W, H)$ denote the image dimensions. Normalized flow vectors are obtained by scaling the horizontal and vertical components by the image width and height, respectively, and the normalized endpoint error is then defined from these vectors:
\begin{equation}
\begin{aligned}
\tilde{\mathbf{F}}(x,y) &= \left( \frac{u(x,y)}{W}, \frac{v(x,y)}{H} \right),
\qquad
\mathrm{EPE}_{norm}(x,y) &= 100\cdot\left\| \tilde{\mathbf{F}}_{pred}(x,y) - \tilde{\mathbf{F}}_{gt}(x,y) \right\|_2.
\end{aligned}
\end{equation}

EPE$_\text{n}$ is thus expressed in percent of the image size. Note that this metric re-weights horizontal versus vertical errors based on the image axes differences.

In addition to full-frame evaluation, we report drone-region metrics that restrict averaging to the image area occupied by the drone. Let $\Omega_D \subseteq \Omega$ denote the set of drone pixels. Then, instead of averaging a per-pixel quantity $m(x,y)$ over $\Omega$, we average only over $\Omega_D$.

\section{The FlyingDrones Dataset}\label{sec:dataset}

\begin{figure*}
    \centering
    \includegraphics[width=\linewidth]{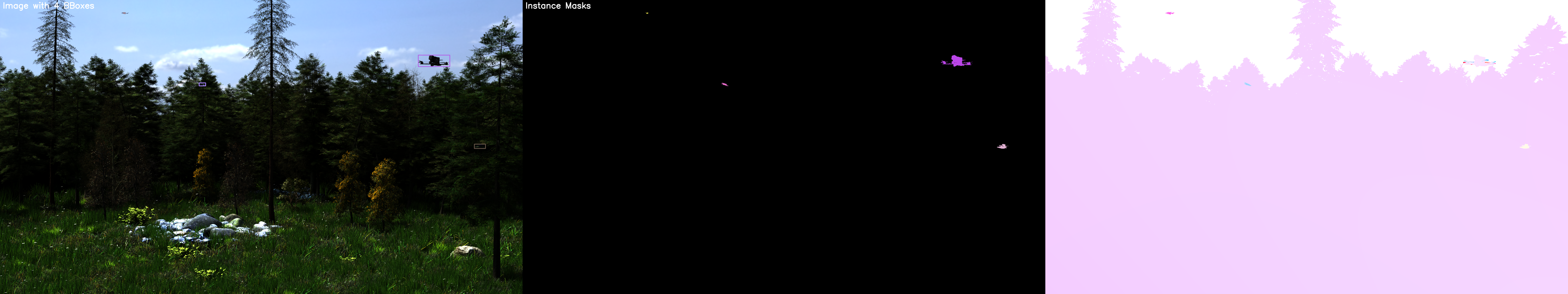}
    
    \vspace*{0.5em}
    
    \includegraphics[width=\linewidth]{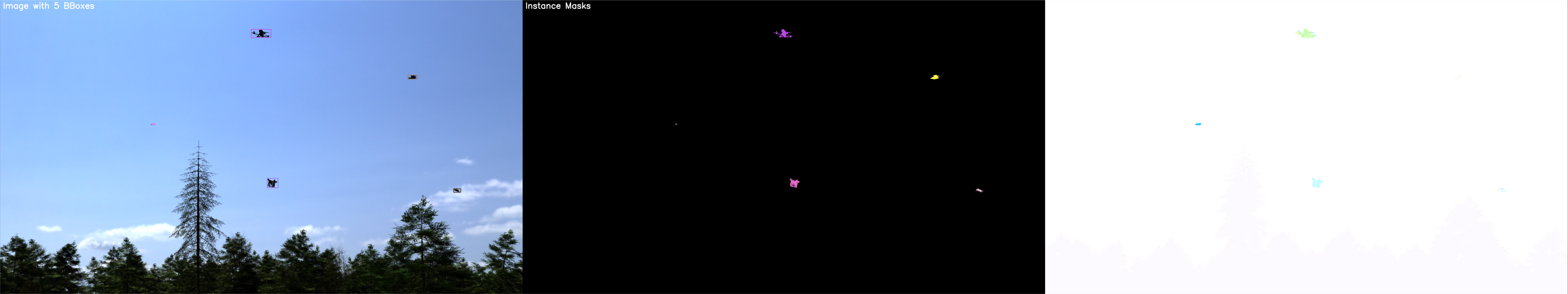}
    \caption{Example images from the dataset. The left column shows the color images with bounding boxes drawn around the drones, the middle column the segmentation masks of the drones, and the right column a color visualization of the optical flow.}
    \label{fig:dataset-examples}
\end{figure*}

Long-range drone footage has no dense optical-flow ground truth, and none can practically be collected at the scale needed for benchmarking, since annotating per-pixel motion for a target that spans a handful of pixels is not a task humans can do reliably by hand. We therefore built a synthetic dataset\footnote{\url{https://www.kaggle.com/datasets/flyingdrones/flyingdrones/data}} in which dense flow and segmentation masks come directly from the renderer rather than from annotation (Figure~\ref{fig:dataset-examples}). All scenes, assets, and animations were assembled in Blender and rendered with the Cycles engine in a headless pipeline that supports parallel generation on GPU servers. Ground-truth annotations are derived directly from the 3D simulation: each drone model is assigned a unique pass index, the renderer exports multi-layer OpenEXR files containing the RGB image, raw Vector pass, and Object Index pass, and a post-processing routine extracts pixel-wise motion vectors and per-drone instance masks consistent with the rendered geometry and lighting. The flow stored for frame $t$ is \emph{backward} flow: it maps every pixel of frame $t$ to its position in frame $t-1$. All evaluations in this paper therefore use the pair $(I_t, I_{t-1})$ with $I_t$ as the reference frame, and temporal models receive the correspondingly time-reversed frame window. Appendix~\ref{sec:real} complements this synthetic evaluation with a real-footage study that establishes which of its findings hold on video from real cameras.

The dataset is released in three resolution variants, all derived from the same native 16:9 render and written as width~$\times$~height (W\,$\times$\,H): \textit{Drones-Large} at $1920\times1080$, the native rendering resolution; \textit{Drones-Standard} at $960\times432$, obtained by resizing \textit{Drones-Large} directly to $960\times432$ (a factor of $0.5$ horizontally and $0.4$ vertically, with the flow components scaled accordingly), which matches the $960\times432$ crop size used as a training and evaluation resolution by several of the benchmarked models (e.g.\ SEA-RAFT's \texttt{TSKH432x960} schedule); and \textit{Drones-Small} at $480\times270$, a four-times downsampled version of \textit{Drones-Large}. Because all three subsets depict the same scenes, performance differences across evaluation resolutions can be attributed directly to spatial resolution rather than to scene content. \textit{Drones-Small} in particular emulates the sampling consequence of long-range observation, reducing drone footprint and apparent motion to fewer pixels without changing the scene. It does not reproduce the other physical consequences of distance such as atmospheric attenuation and turbulence, reduced contrast, and lens or sensor effects, and should be read as a controlled scale ablation rather than a complete long-range simulation (see also the Limitations).

The dataset comprises 9001 frames rendered at the native $1920\times1080$ resolution, yielding 9000 consecutive optical-flow pairs, each indexed by its reference frame $I_t$. Aerial targets are visible in 7908 of the 9000 reference frames. The remaining 1092 are drone-free and excluded from drone-region averages. Scenes contain between zero and six simultaneous drones (2127 frames with one drone, 2448 with two, 1457 with three, 1054 with four, 543 with five, 279 with six), so summing drone instances over all reference frames gives 19\,999. We use \emph{drone instance} for one drone visible in one reference frame, and the drone-region mask $\Omega_D$ used by the metrics is the union of all instance masks in a frame. The instances are drawn from six distinct 3-D multirotor assets, re-used across scenes with varying scale, pose, flight path, and lighting, so appearance diversity comes from viewpoint, range, and illumination rather than from a large asset library.

Table~\ref{tab:dataset_stats} summarizes the size and motion distributions at the native $1920\times1080$ resolution (2 073 600 pixels). The median drone area is 155 pixels, or about $0.0075\%$ of the image area; 25\% of instances occupy 47 pixels or fewer, the minimum target size is a single pixel, and even the largest target (39 275 pixels) occupies less than $1.9\%$ of the image area. Averaged over the drone pixels of each reference frame that contains drones, the ground-truth flow magnitude has a mean of $8.39$ pixels per frame (standard deviation $9.60$, median $5.56$, maximum $111.4$), and individual drone pixels reach $211.2$ pixels per frame, so although the targets are spatially small, their motion ranges from sub-pixel shifts to large displacements.

\begin{table*}[t]
\centering
\caption{FlyingDrones dataset statistics at $1920\times1080$. Drone flow magnitude is the mean ground-truth flow magnitude over the drone pixels of each frame that contains drones (7908 frames), except the maximum, which is over individual drone pixels.}
\label{tab:dataset_stats}
\small
\setlength{\tabcolsep}{4pt}
\begin{minipage}[t]{0.43\textwidth}
\centering
\begin{tabular}{lc}
\toprule
\textbf{General property} & \textbf{Amount} \\
\midrule
Total frames & 9001 \\
Flow pairs & 9000 \\
Frames with visible drones & 7908 \\
Drone instances & 19,999 \\
\midrule
\textbf{Drone flow magnitude} & \textbf{[px/frame]} \\
\midrule
Mean (per frame) & 8.39 \\
Median (per frame) & 5.56 \\
Std (per frame) & 9.60 \\
Max (per pixel) & 211.24 \\
\bottomrule
\end{tabular}
\end{minipage}\hfill
\begin{minipage}[t]{0.53\textwidth}
\centering
\begin{tabular}{lcc}
\toprule
\textbf{Drone size} & \textbf{[px]} & \textbf{[\% of image]} \\
\midrule
Minimum & 1 & 4.8 $\times 10^{-5}$ \\
$p_{25}$ & 47 & 2.3 $\times 10^{-3}$ \\
$p_{50}$ & 155 & 7.5 $\times 10^{-3}$ \\
Mean & 525 & 2.5 $\times 10^{-2}$ \\
$p_{75}$ & 549 & 2.6 $\times 10^{-2}$ \\
$p_{95}$ & 2367 & 1.1 $\times 10^{-1}$ \\
$p_{99}$ & 5194 & 2.5 $\times 10^{-1}$ \\
Maximum & 39275 & 1.9 \\
\bottomrule
\end{tabular}
\end{minipage}
\end{table*}

\section{Results}\label{sec:results-protocol-anchor}

We evaluate 16 architectures (88 public checkpoints in total) on FlyingDrones at all three resolutions, giving 264 checkpoint\,$\times$\,resolution cells, together with a zero-flow reference that predicts $\mathbf{0}$ everywhere and whose drone-region EPE therefore equals the mean ground-truth drone motion. Table~\ref{tab:leaderboard-main} summarizes the best checkpoint of every family; per-resolution leaderboards with confidence intervals are in Tables~\ref{tab:summary-best-drone-270x480}--\ref{tab:summary-best-drone-1080x1920} (Appendix~\ref{sec:appendix-leaderboards}), and checkpoint-level results, model-wise commentary, and training schedules are collected in Appendix~\ref{sec:appendix-model-guide}, Appendix~\ref{sec:appendix-curricula-discussion}, and Table~\ref{tab:checkpoint-schedules}. For reproducibility, Table~\ref{tab:asset_licenses} lists the official source locations and code licenses or public usage terms of the external codebases used in this study.

Evaluating all 88 checkpoints on the full sequence is computationally impractical (several families, notably FlowDiffusion, take 4--18\,s per sample), and a contiguous run of frames would over-represent whichever scenes occur first. We instead evaluate on a fixed stratified subset of 1000 reference frames drawn from the 9000, balanced across drone footprint area (five data-driven quintiles) and drone-region flow magnitude ($<1$, $1$--$3$, $3$--$8$, $\geq8$\,px/frame, plus a drone-free stratum), yielding 21 occupied strata filled by proportional-with-floor sampling so that rare regimes are represented alongside the common mid-range. Frame $0$ has no predecessor under the backward-flow convention, so 999 pairs are evaluated. Selection is keyed by frame index, so the identical pairs are evaluated at all three resolutions and cross-resolution differences remain attributable to spatial resolution alone; because the smallest drones vanish when the frame is downsampled, drone-region averages cover 885, 896 and 898 pairs at $480\times270$, $960\times432$ and $1920\times1080$. All averages are over pairs; 95\% confidence intervals are percentile bootstraps over pairs (10{,}000 resamples). Appendix~\ref{sec:appendix-protocol} compares this sample with a contiguous window of 500 frames: the window is harder for every family, but the family ranking is largely preserved (Spearman $\rho$ between $0.83$ and $0.95$).

Inference ran on a heterogeneous GPU cluster. Eight families (FlowFormer++, GMFlow, LiteFlowNet3, MemFlow, RAFT, UniMatch, VideoFlow, WAFT) were evaluated at $1920\times1080$ with tiled inference (marked $\ddagger$): a grid of $960\times432$ tiles (the \textit{Drones-Standard} crop size) with $20$--$40$\,px minimum overlap, each tile processed by the unmodified upstream model and recomposed by Gaussian-weighted blending (uniform blending for FlowFormer++, GMFlow and UniMatch, whose official evaluators tile natively); FlowFormer++ \texttt{things\_288960} uses its $960\times288$ training crop as the tile. Tiling is applied identically to all checkpoints of a family, but it is a deviation from whole-frame inference and is flagged in every table. No other modification was made to the public model code; only the metric computation is ours.

\subsection{Performance Overview}

\input{tables/tab_leaderboard_main}

Table~\ref{tab:leaderboard-main} reports the best checkpoint of every family at each resolution. Every family beats the zero-flow reference, by $16$--$50\%$ at $480\times270$, $33$--$63\%$ at $960\times432$ and $32$--$79\%$ at $1920\times1080$, and the spread between families grows with resolution: the best-per-family drone-region EPE spans $0.873$--$1.459$ at $480\times270$ ($1.7\times$), $1.293$--$2.336$ at $960\times432$ ($1.8\times$) and $1.557$--$5.125$ at $1920\times1080$ ($3.3\times$). At $1920\times1080$ the families fall into two clear groups: eight global-matching, recurrent, transformer and short-window temporal families stay at or below $2.09$\,px (UniMatch, VideoFlow, FlowFormer++, SEA-RAFT, MemFlow, GMFlow, RAFT, WAFT), while all pyramid, classical, self-supervised and diffusion models are at $3.1$\,px or above. The top ranks are statistically resolved at the two lower resolutions (paired bootstrap: ARFlow beats MemFlow at $480\times270$ and UniMatch beats VideoFlow at $960\times432$ in $\geq99.7\%$ of resamples) but not at $1920\times1080$, where UniMatch and VideoFlow \texttt{BOF\_sintel} are tied ($1.557$ vs.\ $1.582$, UniMatch ahead in $71\%$ of resamples). Within-family spread is comparable to or larger than this between-family spread: at $960\times432$ the leaderboard spans $1.8\times$, while VCN spans $2.3\times$, UniMatch $2.1\times$ and VideoFlow $3.0\times$ across their own checkpoints; at $1920\times1080$ VideoFlow alone spans $13\times$ ($1.58$ to $20.7$, \texttt{MOF\_kitti} collapsing). A family's position is therefore set jointly by its architecture and by which public checkpoint happens to suit this regime --- Appendix~\ref{sec:real} confirms the same effect on real footage.

Full-frame EPE (Appendix~\ref{sec:appendix-model-results}) averages over the whole image and is dominated by background and camera motion, whereas drone-region EPE averages only over ground-truth drone pixels and is the harsher, more relevant score. For UniMatch, which leads at the two higher resolutions, drone-region EPE exceeds full-frame EPE by $3.5\times$ at $480\times270$ ($0.952$ vs.\ $0.275$), $2.3\times$ at $960\times432$ and $1.3\times$ at $1920\times1080$, so full-frame scores understate target-region difficulty most where the target is smallest. The resolution-normalized metric shows the opposite trend to pixel EPE: the best drone-region EPE$_\text{n}$ falls from $0.215$ at $480\times270$ to $0.175$ at $960\times432$ and $0.097$ at $1920\times1080$, i.e.\ the leading models localize the drone's motion more precisely relative to image size as resolution grows, even though their pixel error rises by $1.8\times$ while the drone's own displacement (the zero-flow EPE) rises by $4.3\times$. Pixel-threshold measures make the residual difficulty concrete: the leading models still leave $20$--$27\%$ of drone pixels above $1$\,px error at every resolution and $7$--$12\%$ above $3$\,px, compared with $47$--$86\%$ and $16$--$64\%$ for zero flow.

\subsection{Correlation and Iterative-Refinement Models}\label{sec:corr}

RAFT, SEA-RAFT, and WAFT preserve strong correspondence reasoning without aggressive coarse-to-fine downsampling, but differ in curriculum and matching design: RAFT uses a dense all-pairs correlation volume with recurrent updates over the staged curriculum \texttt{chairs}$\to$\texttt{things}$\to$\texttt{sintel}$\to$\texttt{kitti}; SEA-RAFT simplifies the update and adds uncertainty-aware learning over a longer TartanAir-anchored curriculum; and WAFT uses warping-based alignment with a transformer-style update module and a frozen DepthAnythingV2 encoder~\cite{depth_anything_v2}.

All three families are in the leading cluster at $1920\times1080$ but only mid-field at $480\times270$. SEA-RAFT climbs most with resolution (rank 10, 5 and 4; drone EPE $1.117$, $1.557$, $1.883$ for its Sintel release), consistent with its TartanAir-anchored, $960\times432$-crop curriculum. RAFT ranks 11, 8 and 7 ($1.152$, $1.713$, $2.047$); its Sintel checkpoint is best at the two higher resolutions and its spread grows from $1.2\times$ to $1.9\times$, so the last curriculum stage matters most where displacements are largest. WAFT ranks 6, 9 and 8, with its Spring-540p release best at the two lower resolutions and the \texttt{tar-c-t} base best at $1920\times1080$. The three released SEA-RAFT checkpoints differ by up to $1.6\times$ at $1920\times1080$ (Sintel $1.883$, KITTI $2.026$, Spring $2.972$), reinforcing the checkpoint-dominance observation.

\subsection{Classical, Pyramid, and Early Deep Baselines}\label{sec:classical}

FlowNet2, PWC-Net, LiteFlowNet3, VCN, and PyFlow cover the classical coarse-to-fine and early deep-learning baselines, all relying on hierarchical, coarse-to-fine estimation. This is effective on standard benchmarks but prone to attenuating a target that is only a tiny image structure. PyFlow is a hand-crafted variational method with no learned pre-training; FlowNet2 stacks multiple subnetworks trained in stages; PWC-Net and LiteFlowNet3 are compact pyramid descendants; VCN adds volumetric correspondence reasoning. At $480\times270$, where about half of the drone pixels move by less than one pixel, these models are competitive: PyFlow is fourth overall ($0.977$) and VCN seventh ($1.067$). As resolution grows they fall behind, and at $1920\times1080$ all five form the lower group ($3.14$--$4.40$, against $1.56$--$2.09$ for the leading cluster), consistent with coarse-to-fine estimation losing the small target once its displacement exceeds the coarse levels' resolution. VCN, which unlike the tiled LiteFlowNet3 runs on the whole $1920\times1080$ frame, also has the second-largest within-family spread there ($8.1\times$). PyFlow nevertheless remains a strong real-footage baseline (Appendix~\ref{sec:real}).

\subsection{Efficient and Reduced-Supervision Models}\label{sec:efficient}

ARFlow reduces dependence on dense labels through self-supervised consistency objectives; FastFlowNet keeps supervised training but uses a lightweight pyramid architecture for speed. ARFlow's KITTI12 multi-view checkpoint is the best model overall at $480\times270$ ($0.873$) and also has by far the lowest full-frame EPE there ($0.150$ against $\geq0.234$ for every other family). An independent check attributes the full-frame advantage to near-zero output on the static background, which covers about two thirds of the pixels at this resolution, while on moving background it matches RAFT; its photometric self-supervision therefore acts as a low-noise prior rather than as a more accurate motion estimator. ARFlow drops to rank 6 at $960\times432$ and 10 at $1920\times1080$, and FastFlowNet is last or second to last at every resolution ($1.459$, $2.336$, $4.425$), so lightweight pyramids do not scale to the high-resolution regime.

\subsection{Transformer and Global-Matching Models}\label{sec:transformer}

GMFlow, UniMatch, and FlowFormer++ treat optical flow as a dense correspondence problem with broader spatial context, via global attention or transformer-style decoding rather than local cost volumes and recurrent corrections. UniMatch is the best family at $960\times432$ and $1920\times1080$ ($1.293$, $1.557$) and third at $480\times270$ ($0.952$), and it also attains the lowest drone-region EPE$_\text{n}$ at the two higher resolutions. FlowFormer++ ranks 9, 4 and 3, and GMFlow 15, 7 and 6. This family is also a clear illustration of curriculum dominance: FlowFormer++ spans $1.8\times$ at $960\times432$ and $1.9\times$ at $1920\times1080$ across four checkpoints, and GMFlow $2.2\times$ at $1920\times1080$.

\subsection{Temporal, Memory-Based, and Generative Models}\label{sec:temporal}

VideoFlow uses multi-frame input windows optimized for temporally coherent video flow; MemFlow adds an explicit memory module on top of recurrent flow estimation; FlowDiffusion formulates prediction as a diffusion-style generative process. All temporal models receive genuine neighbouring frames. VideoFlow's three-frame \texttt{BOF\_sintel} is second at $960\times432$ and $1920\times1080$ ($1.425$, $1.582$; tied with UniMatch at the latter), but its five-frame MOF variants do not improve on it: BOF beats MOF for 11 of the 12 matched checkpoint\,$\times$\,resolution pairs, and \texttt{MOF\_kitti} collapses at $1920\times1080$ ($20.7$). MemFlow, which carries a memory over a five-frame clip, is second at $480\times270$ ($0.948$, Spring checkpoint) and stays in the leading cluster at every resolution. FlowDiffusion is mid-field at $480\times270$ ($1.099$) but falls to the bottom at the two higher resolutions ($2.220$, $5.125$). A short temporal window is therefore competitive, but neither longer windows nor generative refinement add accuracy on synthetic drone motion.

\section{Discussion}

The results support optical flow as a useful cue for detecting small aerial targets: every evaluated family recovers drone motion clearly better than the zero-flow reference, and the leading models keep their drone-region error between roughly $0.9$ and $1.6$\,px while the drone's own displacement grows fourfold from $480\times270$ to $1920\times1080$. The divergence between pixel-based and resolution-normalized metrics, and between full-frame and drone-region scores, shows why scale-aware reporting is necessary. Rankings change with resolution (Table~\ref{tab:leaderboard-main}): self-supervised (ARFlow), memory-based (MemFlow), global-matching (UniMatch) and even classical (PyFlow) models lead at $480\times270$, where about half of the drone pixels move by less than one pixel, whereas global-matching and short-window temporal models (UniMatch, VideoFlow BOF) lead at the two higher resolutions, and pyramid, classical and diffusion models fall clearly behind at $1920\times1080$.

Aggregating the appendix tables over the twelve families that release both a KITTI- and a Sintel-fine-tuned checkpoint, Sintel fine-tuning is better for 8 of 12 families at $480\times270$ and 9 of 12 at both higher resolutions, and its median relative advantage grows with resolution ($+3.5\%$, $+20\%$, $+22\%$). The exceptions are ARFlow, FastFlowNet and WAFT, for which the KITTI release is better at every resolution (and VCN at $480\times270$, by less than $1\%$). Displacement range does not explain this pattern (Appendix~\ref{sec:appendix-datasets}): the labelled pixels of the KITTI 2015 training set move more than Sintel's (median $23$ vs.\ $3.9$\,px), and the drone motion lies between the two (per-frame median $5.56$\,px at $1920\times1080$, Table~\ref{tab:dataset_stats}). The cause remains open; dense synthetic versus sparse LiDAR supervision and scene content are candidate factors. The practical magnitude of this curriculum effect is large: within-family swings between the best and worst public checkpoint reach $2.1$--$3.0\times$ at $960\times432$ and up to $13\times$ at $1920\times1080$, comparable to or larger than the $1.8\times$ and $3.3\times$ spans of the entire best-per-family leaderboard. Curriculum selection, not architecture selection alone, is therefore a dominant lever for drone-region accuracy and Appendix~\ref{sec:real} shows the same effect on real footage.

Two appendix studies connect the synthetic results to real data. On 21 cameras of real UAV footage (Appendix~\ref{sec:real}), synthetic drone-region EPE and real drone localization are positively rank-correlated over 41 checkpoints (Spearman $\rho=0.54$--$0.63$ across the three synthetic resolutions, $p<10^{-3}$), although the exact top ranks differ: MemFlow leads on real footage and is in the synthetic top five at every resolution, whereas classical PyFlow is much stronger and VideoFlow's five-frame variant much weaker on real footage than on synthetic data. Fine-tuning on 800 FlyingDrones pairs (Table~\ref{tab:ft-main}, Appendix~\ref{sec:finetune}) reduces held-out drone-region EPE by $19$--$37\%$ for three of four architectures and improves real-footage localization for all four, most strongly for VideoFlow MOF, whose public checkpoint transfers worst.

\input{tables/tab_ft_main}

For practical aerial surveillance, the choice of public checkpoint matters as much as the choice of architecture, and the best choice depends on the operating resolution. Future work should prioritize native-resolution aerial training data and evaluate candidate checkpoints at the deployment resolution; both pixel-space and normalized metrics remain necessary, and no current model reaches sub-pixel drone-region EPE at $960\times432$ or $1920\times1080$.

\subsection*{Limitations}
The benchmark itself is synthetic and although the renderer provides dense ground-truth flow at very small target scales, it cannot capture all field effects relevant to long-range aerial sensing, including atmospheric turbulence, sensor noise, compression artifacts, rolling-shutter distortions, and platform-specific optics, and \textit{Drones-Small} therefore emulates the sampling consequence of range, not its full radiometry. The real-footage study of Appendix~\ref{sec:real} mitigates but does not remove this concern: it uses static-camera sequences with point (not dense) labels, so it measures motion localisation and agreement at the target rather than full-field accuracy, and its absolute numbers are a proxy rather than an EPE; it is evaluated across 41 real-footage checkpoints and 21 cameras, but moving platforms, multiple simultaneous targets, and additional sensors and scenes remain future work. The evaluation resolution ceiling is $1920\times1080$, the native rendering resolution of the current release; because rankings and the best curriculum both shift with resolution, extrapolating this ordering to 2K or 4K sensors is not warranted, and re-rendering the retained scene geometry and flight paths at higher resolution is the natural, if costly, extension.

Because the main study compares publicly available checkpoints with different curricula, the results reflect the joint effect of training domain and architecture rather than a controlled ablation over optimization budget, model size, and compute. We do not resolve this confound, retraining 16 architectures under one curriculum is beyond our compute budget, but we bound it by reporting all 88 public checkpoints so that within-family spread is visible (Appendix~\ref{sec:appendix-model-results}), and by the controlled fine-tuning study of Appendix~\ref{sec:finetune}, in which four architectures are trained from a common in-domain split under a matched recipe. That study shows gains on held-out FlyingDrones data and on the static-camera real footage of Appendix~\ref{sec:real}, but it covers four architectures and one training budget only.

Finally, the benchmark isolates optical-flow quality from downstream detection and tracking; strong flow estimates for tiny targets may still be insufficient in practice, where false positives, latency, and calibration errors also matter.

\subsection*{Broader Impact}
Improved motion estimation for small aerial targets may support applications such as airspace safety and wildlife-aware collision avoidance. This work contributes a benchmark, not a deployable surveillance or tracking system. Because drone-analysis data can still be used in surveillance settings, we recommend application-specific validation, human oversight, and clearly documented operational limits for downstream deployment.

\section{Conclusion}
We introduce FlyingDrones, a synthetic benchmark for optical flow on drone motion, together with drone-region metrics and resolution-normalized EPE for cross-resolution comparison. Such motion is characterized by very small motion regions, a case in which state-of-the-art methods still leave substantial error.
Across 16 optical-flow families and 88 public checkpoints, evaluated on a motion- and size-stratified subset of 999 flow pairs, every family recovers drone motion better than a zero-flow reference, and the separation between families grows with resolution: at $1920\times1080$, UniMatch ($1.557$\,px drone-region EPE) and VideoFlow's three-frame variant ($1.582$\,px) lead a cluster of global-matching, recurrent, transformer and short-window temporal models, while pyramid, classical, self-supervised, lightweight, and diffusion models remain at $3.1$\,px or above. At $480\times270$, where about half of the drone pixels move by less than one pixel, the self-supervised ARFlow ($0.873$\,px), MemFlow, UniMatch, and even the classical PyFlow lead. Training data and curriculum matter as much as architecture: within-family spread between public checkpoints reaches $2$--$3\times$ at $960\times432$ and up to $13\times$ at $1920\times1080$, and Sintel-style fine-tuning increasingly outperforms KITTI-style fine-tuning as resolution grows; a five-frame temporal window adds nothing over three frames.
Two complementary studies test the benchmark beyond synthetic accuracy. A ground-truth-free motion-alignment evaluation on 21 real UAV cameras confirms the regime-level conclusions, sub-pixel motion is the binding constraint and drone-region evaluation is necessary, and shows that the synthetic ranking correlates moderately with real-world drone localization (Spearman $\rho=0.63$), while the exact top ranks differ, so FlyingDrones is best used to shortlist architectures and training domains before in-domain validation. A fine-tuning study on a held-out FlyingDrones split shows that 800 in-domain pairs reduce drone-region error by $19$--$37\%$ for three of four architectures and improve drone localization on real footage for all four (VideoFlow $\mathrm{NSS}_D$ $4.9\rightarrow17.8$).
Overall, the benchmark points to three priorities for small-aerial-target flow: scale-aware, drone-region evaluation at the deployment resolution, curriculum selection validated on in-domain footage, and in-domain training data, for which high-resolution aerial data remains the main missing part.

\begin{ack} 
This article is the result of a collaboration with the partners of the STORE project (\url{https://edf-store.com}), funded
by the European Defence Fund (EDF) under grant agreement EDF-2022-101121405-STORE and supported by the Munich Center for Machine Learning (MCML).
\end{ack}

\bibliographystyle{IEEEtranN}
\bibliography{references}

\FloatBarrier

\appendix

\section{Licenses of External Assets}\label{sec:appendix-licenses}
Table~\ref{tab:asset_licenses} summarizes the licenses or public terms currently associated with the external datasets, benchmark assets, and model repositories referenced in this work. We use these assets only for evaluation and discussion of published checkpoints. We do not redistribute third-party datasets or model weights as part of this submission. Where the public benchmark page does not expose a clear standalone license, we state that explicitly.

\begingroup
\setlength{\tabcolsep}{4pt}
\scriptsize
\renewcommand{\arraystretch}{1.08}
\begin{longtable}{>{\raggedright\arraybackslash}p{2.5cm}>{\raggedright\arraybackslash}p{1.5cm}>{\raggedright\arraybackslash}p{2.7cm}>{\raggedright\arraybackslash}p{5.1cm}}
\caption{Licenses of external datasets and model repositories referenced in this work.}
\label{tab:asset_licenses}\\
\toprule
\textbf{Asset} & \textbf{Type} & \textbf{Official source} & \textbf{License / terms} \\
\midrule
\endfirsthead

\multicolumn{4}{l}{\textit{Table~\ref{tab:asset_licenses} continued from previous page.}} \\
\toprule
\textbf{Asset} & \textbf{Type} & \textbf{Official source} & \textbf{License / terms} \\
\midrule
\endhead

\midrule
\multicolumn{4}{r}{\textit{Continued on next page.}} \\
\endfoot

\bottomrule
\endlastfoot

FlyingChairs & Dataset & University of Freiburg~\cite{dosovitskiy2015} & Non-commercial / research use only; no standard open-data license stated. \\
FlyingThings3D & Dataset & University of Freiburg Scene Flow Datasets~\cite{mayer_large_2016} & Non-commercial / research use only; no standard open-data license stated. \\
MPI Sintel & Dataset / source movie assets & MPI Sintel benchmark; Blender Durian project~\cite{Butler:ECCV:2012} & Derived from the open movie \textit{Sintel}; movie and published project assets are released under CC BY 3.0; dataset published under PS:License 1.0. \\
KITTI & Dataset & KITTI Vision Benchmark Suite~\cite{Menze2015CVPR} & CC BY-NC-SA 3.0. \\
TartanAir & Dataset & TartanAir documentation~\cite{tartanair2020iros} & CC BY 4.0 for the dataset; accompanying toolkit under MIT. \\
Spring & Dataset & Spring benchmark / DaRUS release~\cite{mehl2023} & CC BY 4.0. \\
HD1K & Dataset & HD1K benchmark suite~\cite{kondermann2016} & No standalone public license found; download distributed through the benchmark site with protected access. \\
FlowNet2 & Model code & \url{https://github.com/lmb-freiburg/flownet2} & BSD 2-Clause (Caffe-derived); however, the repository documentation explicitly restricts use to non-commercial research only and requires separate author consent for commercial use. \\
PyFlow & Model code & \url{https://github.com/pathak22/pyflow} & MIT (Python wrapper only; Copyright \textcopyright\ 2017 Deepak Pathak). The underlying Ce Liu optical flow C++ implementation carries separate terms; consult \url{https://people.csail.mit.edu/celiu/OpticalFlow/} for the base library license. \\
PWC-Net & Model code & \url{https://github.com/NVlabs/PWC-Net} & CC BY-NC-SA 4.0 (Copyright \textcopyright\ 2018 NVIDIA Corporation); non-commercial use only with share-alike and attribution requirements; commercial use requires separate permission from NVIDIA. \\
LiteFlowNet3 & Model code & \url{https://github.com/twhui/LiteFlowNet3} & Custom non-commercial license; software and paper assets are provided for non-commercial use only without any warranty; commercial use requires a prior arrangement with the author (Tak-Wai Hui). \\
VCN & Model code & \url{https://github.com/gengshan-y/VCN} & MIT. \\
ARFlow & Model code & \url{https://github.com/lliuz/ARFlow} & MIT. \\
FastFlowNet & Model code & \url{https://github.com/ltkong218/FastFlowNet} & MIT. \\
RAFT & Model code & \url{https://github.com/princeton-vl/RAFT} & BSD 3-Clause. \\
GMFlow & Model code & \url{https://github.com/haofeixu/gmflow} & Apache 2.0. \\
UniMatch & Model code & \url{https://github.com/autonomousvision/unimatch} & MIT. \\
FlowFormer++ & Model code & \url{https://github.com/XiaoyuShi97/FlowFormerPlusPlus} & Apache 2.0. \\
VideoFlow & Model code & \url{https://github.com/XiaoyuShi97/VideoFlow} & Apache 2.0. \\
FlowDiffusion (DDVM) & Model code & \url{https://github.com/google-research/google-research} & Apache 2.0 (monorepo-level license covering all sub-projects including DDVM). \\
MemFlow & Model code & \url{https://github.com/DQiaole/MemFlow} & Apache 2.0. \\
SEA-RAFT & Model code & \url{https://github.com/princeton-vl/SEA-RAFT} & BSD 3-Clause. \\
WAFT & Model code & \url{https://github.com/princeton-vl/WAFT} & BSD 3-Clause. \\
DepthAnythingV2 encoder & Model code / pretrained weights & \url{https://github.com/DepthAnything/Depth-Anything-V2} & Code under Apache 2.0; model weights are mixed-license: Small variant under Apache 2.0, Base/Large/Giant variants under CC BY-NC 4.0. \\
\end{longtable}
\endgroup

Our FlyingDrones dataset is publicly available on Kaggle at \url{https://www.kaggle.com/datasets/flyingdrones/flyingdrones/data} together with documentation and release limitations. The models are run through their official third-party repositories (Table~\ref{tab:asset_licenses}). We release the dataset loader, which encodes the backward-flow pairing convention, the implementation of the normalized and drone-specific metrics, the stratified evaluation and fine-tuning splits, and the per-model evaluation wrappers in a public code repository linked from the dataset page.

\section{Checkpoint Specific Model Results}

\subsection{How to read the checkpoint tables}
\label{sec:appendix-model-guide}

The appendix reports every checkpoint in two aggregated tables, a full-frame evaluation (Table~\ref{tab:agg-full}) and a drone-pixel-only evaluation (Table~\ref{tab:agg-drone}), grouped by model family, resolution and checkpoint. For the task studied in this paper the drone-pixel-only table is the decisive one, because distant UAVs occupy only a tiny fraction of the image and their motion can easily be obscured by the much larger background flow generated by camera motion. The checkpoint with the lowest full-frame EPE at every resolution (ARFlow \texttt{KITTI12/pwclite\_ar\_mv}) is first on the drone region only at $480\times270$. Both tables end with the zero-flow reference, whose drone-region EPE equals the mean ground-truth drone motion ($1.744$, $3.498$ and $7.549$\,px at the three resolutions); a checkpoint at or above this value recovers no usable drone motion. Two checkpoints reach this level at $960\times432$ and eight at $1920\times1080$, all of them outside the leaderboards. Runs marked $\ddagger$ used tiled inference at $1920\times1080$ (Section~\ref{sec:results-protocol-anchor}); VCN has no tiling option in its evaluator and is run on the full $1920\times1080$ frame.

Two release-level details matter when reading the tables. GMFlow's \texttt{gmflow\_things}, \texttt{gmflow\_with\_refine\_things} and \texttt{gmflow\_with\_refine\_sintel} are the same released weights as UniMatch's \texttt{gmflow-scale1-things}, \texttt{gmflow-scale2-things} and \texttt{gmflow-scale2-sintel}, and the paired rows agree up to floating-point noise; we keep both rows so that each family's table is complete. SEA-RAFT, in contrast, releases three distinct files (\texttt{Tartan-C-T-TSKH-kitti432x960-M}, the Sintel release \texttt{Tartan-C-T-TSKH432x960-M}, and \texttt{Tartan-C-T-TSKH-spring540x960-M}), which score differently. We use the public checkpoint names and release descriptions whenever those schedules are exposed directly in Table~\ref{tab:checkpoint-schedules}. If the full optimization history is not visible from the release, we report only the most conservative curriculum that is supported by the published checkpoint naming.

\begingroup
\setlength{\LTleft}{0pt}
\setlength{\LTright}{0pt}
\scriptsize
\renewcommand{\arraystretch}{1.08}
\begin{longtable}{p{0.13\linewidth}p{0.30\linewidth}p{0.47\linewidth}}
\caption{Training schedules for the checkpoint families used in the appendix result tables. Arrows indicate the order of datasets used during pre-training and fine-tuning. ``Release snapshot'' denotes checkpoints whose public names expose optimization stage but not a fully disaggregated dataset-by-dataset curriculum.}\label{tab:checkpoint-schedules}\\
\toprule
Family & Checkpoint(s) & Dataset order used in training \\
\midrule
\endfirsthead
\toprule
Family & Checkpoint(s) & Dataset order used in training \\
\midrule
\endhead
FlowNet2 & \texttt{FlowNet2-C\_checkpoint} & FlyingChairs \\
FlowNet2 & \texttt{FlowNet2-CS\_checkpoint} & FlyingChairs $\rightarrow$ FlyingThings3D \\
FlowNet2 & \texttt{FlowNet2-CSS-ft-sd\_checkpoint}, \texttt{FlowNet2\_checkpoint} & FlyingChairs $\rightarrow$ FlyingThings3D $\rightarrow$ ChairsSDHom / small-displacement refinement \\
FlowNet2 & \texttt{FlowNet2-S\_checkpoint}, \texttt{FlowNet2-SD\_checkpoint} & Published single-stream / small-displacement release checkpoints \\
PyFlow & \texttt{accurate}, \texttt{default}, \texttt{fast} & No learned schedule; classical coarse-to-fine optimization \\
PWC-Net & \texttt{pwc\_net\_chairs}, \texttt{pwc\_net} & FlyingChairs $\rightarrow$ published PWC-Net release refinement \\
LiteFlowNet3 & \texttt{liteflownet3s-sintel}, \texttt{liteflownet3-sintel} & Synthetic pre-training $\rightarrow$ Sintel fine-tuning \\
LiteFlowNet3 & \texttt{liteflownet3s-kitti}, \texttt{liteflownet3-kitti} & Synthetic pre-training $\rightarrow$ KITTI fine-tuning \\
RAFT & \texttt{raft-chairs} & FlyingChairs \\
RAFT & \texttt{raft-things}, \texttt{raft-small} & FlyingChairs $\rightarrow$ FlyingThings3D \\
RAFT & \texttt{raft-sintel} & FlyingChairs $\rightarrow$ FlyingThings3D $\rightarrow$ Sintel \\
RAFT & \texttt{raft-kitti} & FlyingChairs $\rightarrow$ FlyingThings3D $\rightarrow$ Sintel $\rightarrow$ KITTI \\
VCN & \texttt{.../chairs/...}, \texttt{.../things/...} & FlyingChairs or FlyingThings3D \\
VCN & \texttt{.../sintel-ft-trainval/...}, \texttt{.../kitti-ft-trainval/...} & FlyingThings3D $\rightarrow$ Sintel or KITTI fine-tuning \\
ARFlow & \texttt{CityScapes/...}, \texttt{KITTI12/...}, \texttt{KITTI15/...}, \texttt{Sintel/...} & Self-supervised training on the named target dataset \\
FastFlowNet & \texttt{fastflownet\_chairs}, \texttt{fastflownet\_things3d} & FlyingChairs $\rightarrow$ FlyingThings3D \\
FastFlowNet & \texttt{fastflownet\_ft\_sintel}, \texttt{fastflownet\_ft\_kitti}, \texttt{fastflownet\_ft\_mix} & Synthetic pre-training $\rightarrow$ Sintel, KITTI, or mixed fine-tuning \\
GMFlow & \texttt{...chairs...}, \texttt{...things...} & FlyingChairs or FlyingThings3D release checkpoints \\
GMFlow & \texttt{...sintel...}, \texttt{...kitti...}, \texttt{...with\_refine...} & Base checkpoint $\rightarrow$ Sintel or KITTI refinement, optionally with a refinement head \\
UniMatch & \texttt{...things...} & FlyingThings3D \\
UniMatch & \texttt{...mixdata...} & FlyingThings3D $\rightarrow$ mixed-data training \\
UniMatch & \texttt{...sintelft...}, \texttt{...kitti15...} & Things / mixdata $\rightarrow$ Sintel or KITTI fine-tuning \\
FlowFormer++ & \texttt{things}, \texttt{things\_288960} & FlyingThings3D \\
FlowFormer++ & \texttt{sintel}, \texttt{kitti} & FlyingThings3D $\rightarrow$ Sintel or KITTI fine-tuning \\
VideoFlow & \texttt{BOF\_*}, \texttt{MOF\_*} & Synthetic video flow pre-training $\rightarrow$ named fine-tuning set \\
FlowDiffusion & \texttt{pipeline-305000}, \texttt{pipeline-900000} & Release snapshots from the published diffusion pipeline \\
MemFlow & \texttt{MemFlowNet\_things}, \texttt{MemFlowNet\_T\_things} & FlyingThings3D \\
MemFlow & \texttt{MemFlowNet\_spring}, \texttt{MemFlowNet\_sintel}, \texttt{MemFlowNet\_kitti} and \texttt{MemFlowNet\_T\_*} variants & FlyingThings3D $\rightarrow$ Spring, Sintel, or KITTI fine-tuning \\
MemFlow & \texttt{MemFlowNet\_T\_things\_kitti} & FlyingThings3D $\rightarrow$ KITTI \\
SEA-RAFT & \texttt{Tartan-C-T-TSKH432x960-M} (Sintel), \texttt{Tartan-C-T-TSKH-kitti432x960-M}, \texttt{Tartan-C-T-TSKH-spring540x960-M} & TartanAir $\rightarrow$ FlyingChairs $\rightarrow$ FlyingThings3D $\rightarrow$ TSKH mixture (Things, Sintel, KITTI, HD1K), then the named benchmark fine-tune \\
WAFT & \texttt{chairs-things} (implicit base) & FlyingChairs $\rightarrow$ FlyingThings3D, with frozen DepthAnythingV2 encoder \\
WAFT & \texttt{tar}, \texttt{tar-c}, \texttt{tar-c-t} & chairs-things $\rightarrow$ TartanAir-style refinement $\rightarrow$ optional extra curriculum stages \\
WAFT & \texttt{tar-c-t-sintel}, \texttt{tar-c-t-kitti}, \texttt{tar-c-t-spring-540p}, \texttt{tar-c-t-spring-1080p} & chairs-things $\rightarrow$ TartanAir-style refinement $\rightarrow$ Sintel, KITTI, or Spring fine-tuning \\
\bottomrule
\end{longtable}
\endgroup

\subsection{Best-Checkpoint Leaderboards}
\label{sec:appendix-leaderboards}

Tables~\ref{tab:summary-best-drone-270x480}--\ref{tab:summary-best-drone-1080x1920} report one row per model family and resolution, selecting the checkpoint with the lowest drone-region EPE, with 95\% bootstrap confidence intervals over pairs and the zero-flow reference as the last row. Neighbouring entries often have overlapping intervals; a paired bootstrap over the same frames resolves the first rank at $480\times270$ (ARFlow first in $99\%$ of resamples) and at $960\times432$ (UniMatch, $>99\%$), but not at $1920\times1080$, where UniMatch and VideoFlow \texttt{BOF\_sintel} share it ($71\%$ and $29\%$). Re-selecting each family's best checkpoint inside every resample changes these probabilities by at most one percentage point. Section~\ref{sec:results-protocol-anchor} discusses the patterns visible across the three tables.

\input{tables/tab_summary_best_drone_270x480}

\input{tables/tab_summary_best_drone_432x960}

\input{tables/tab_summary_best_drone_1080x1920}

\subsection{Model Specific Evaluation Results}
\label{sec:appendix-model-results}

\input{tables/tab_agg_full}

\input{tables/tab_agg_drone}

\subsection{Pre-Training Datasets and Their Motion Distributions}
\label{sec:appendix-datasets}

Because the apparent motion of a distant drone is statistically insignificant, often sub-pixel to a few pixels, when compared to much larger ego-motion-induced background flow, the choice of pre-training and fine-tuning datasets shapes results far more than architecture alone. Before discussing the curricula, we summarise the motion statistics and scene types of the datasets that recur in the curricula of all 16 families. The distribution of drone target footprints in our own benchmark (Table~\ref{tab:dataset_stats}), median area $155$\,px ($0.0075\%$ of the image) and mean drone flow magnitude $8.39$\,px/frame (median $5.56$) at 1920$\times$1080, is the reference point against which each pre-training dataset should be read.

\subsubsection{FlyingChairs~\cite{dosovitskiy2015}}
22{,}872 image pairs at 512$\times$384 of textured 3D chair models translated and rotated in the image plane over random Flickr backgrounds. Motion is purely planar (2D affine), occlusions are limited, and the displacement distribution is moderate, with a mean magnitude of approximately $11$\,px. There are no real 3D parallax cues. FlyingChairs is the most widely used initial pre-training stage in the field and appears in essentially every supervised curriculum we evaluate (FlowNet2, PWC-Net, LiteFlowNet3, FastFlowNet, RAFT, VCN, GMFlow, UniMatch, FlowFormer++, MemFlow, VideoFlow, SEA-RAFT, WAFT). Because every modern model has seen Chairs, performance differences in our benchmark cannot be attributed to its absence; rather, Chairs sets a common floor on top of which the later, more discriminating fine-tunes operate. Chairs alone contains no representative aerial ego-motion, no atmospheric effects, and no genuinely small targets at long range, and checkpoints that stop at the Chairs stage are accordingly weak: \texttt{raft-chairs}, \texttt{FlowNet2-C} and \texttt{gmflow\_chairs} trail the family best at every resolution (e.g.\ $3.280$, $3.841$ and $4.367$ drone EPE at $1920\times1080$, against $2.047$, $3.136$ and $1.996$ for the respective family best). The one Chairs-only family winner, VCN \texttt{chairs} at $1920\times1080$ ($3.829$), wins only because VCN's later-stage checkpoints break down on the untiled full frame (Appendix~\ref{sec:appendix-curricula-discussion}).

\subsubsection{FlyingThings3D~\cite{mayer_large_2016}}
About 22{,}000 frames at 960$\times$540 of randomly textured 3D objects flying through procedurally generated scenes. Motions are deliberately aggressive, with maximum displacements of several hundred pixels being common, and many pixels see large rotations and out-of-plane translations. This is the canonical heavy-displacement synthetic stage in almost every modern flow curriculum (RAFT, GMFlow, FlowFormer++, MemFlow, VideoFlow, WAFT, FlowNet2, FastFlowNet, VCN). Its very large motions teach robust correlation matching but provide essentially no examples of the sub-pixel residual motion that defines distant drones in our benchmark. Even so, a checkpoint that stops at the Things stage is often competitive: FlowFormer++'s Things releases are the family best at every resolution (third overall at $1920\times1080$, $1.840$), UniMatch's \texttt{regrefine6-things} leads the entire benchmark at $960\times432$ ($1.293$), and Things checkpoints are the family best for RAFT, VCN, GMFlow and VideoFlow at $480\times270$.

\subsubsection{ChairsSDHom~\cite{ilg2017}}
A small-displacement variant introduced for FlowNet2: chairs are placed over more homogeneous, less-textured backgrounds and translate by very small amounts (sub-pixel to a few pixels). It is the only standard pre-training dataset whose motion distribution directly resembles the regime relevant for distant aerial targets. Within the FlowNet2 release, the \texttt{CSS-ft-sd}, \texttt{SD}, and full \texttt{FlowNet2} ensemble checkpoints all incorporate ChairsSDHom-style refinement. The stand-alone small-displacement network \texttt{SD} is worse than the Chairs$\rightarrow$Things \texttt{CS} checkpoint at every resolution ($1.263$/$2.237$/$4.162$ versus $1.226$/$2.011$/$3.217$), and the full \texttt{FlowNet2} ensemble, which fuses the small-displacement branch with the large-displacement stack, improves on \texttt{CS} by at most $0.08$\,px (it is the family best at $480\times270$ and $1920\times1080$, $1.209$ and $3.136$, while \texttt{CS} is best at $960\times432$). Matching displacement statistics alone is therefore insufficient, domain-realistic backgrounds and camera motion also matter.

\subsubsection{MPI Sintel~\cite{Butler:ECCV:2012}}
1{,}064 frames at 1024$\times$436 from the open-source animated film \emph{Sintel}, with Clean and Final passes (the latter adds motion blur, atmospheric effects, and defocus). Its per-pixel flow magnitude has a median of $3.9$\,px and a mean of $13.4$\,px: $21\%$ of the pixels move by less than $1$\,px, and the tail extends past $100$\,px ($1.6\%$ of the pixels) due to fast camera motion and non-rigid characters (measured on every fourth training flow field). Sintel is, alongside KITTI, the canonical single-domain fine-tune in the field: it appears as a dedicated fine-tune in RAFT, LiteFlowNet3, FastFlowNet, GMFlow, UniMatch, FlowFormer++, MemFlow, VideoFlow, WAFT, ARFlow, and as part of the SEA-RAFT mixed (TSKH) refinement; VCN exposes \texttt{sintel-ft-trainval} as a corresponding checkpoint. This near-universal coverage makes Sintel the cleanest dataset for cross-architecture comparison. In our benchmark it is also the most reliable final fine-tune: across the 12 families that release both a Sintel and a KITTI checkpoint, the better Sintel checkpoint beats the better KITTI checkpoint in 8, 9 and 9 families at $480\times270$, $960\times432$ and $1920\times1080$, and the median relative reduction in drone-region EPE grows with resolution ($3.5\%$, $20.0\%$, $22.0\%$). At $1920\times1080$, six of the eight families in the leading cluster ($\leq 2.09$\,px) reach it with a Sintel-finetuned checkpoint, including the two leaders, UniMatch \texttt{regrefine6-sintelft} ($1.557$) and VideoFlow \texttt{BOF\_sintel} ($1.582$). The exceptions are the self-supervised ARFlow, the lightweight FastFlowNet and WAFT, for which KITTI wins at every resolution, and VCN at $480\times270$, where the two differ by less than $1\%$.

\subsubsection{KITTI 2012/2015~\cite{Menze2015CVPR}}
Real driving scenes captured at $1242\times 375$ in Karlsruhe, with sparse LiDAR-derived ground truth that covers mainly the road, vehicles and nearby structures. Over the labelled pixels of the 200 KITTI~2015 training pairs, displacements are large and mostly horizontal, dominated by forward camera motion: median $23$\,px ($18$\,px for non-occluded pixels; median horizontal component $19$\,px), only $9\%$ of the pixels move by less than $1$\,px, and $9\%$ by more than $100$\,px. KITTI fine-tuning is, like Sintel, near-universal: dedicated KITTI checkpoints exist for RAFT, LiteFlowNet3, FastFlowNet, GMFlow, UniMatch, FlowFormer++, MemFlow, VideoFlow, WAFT, ARFlow (KITTI12 and KITTI15), and VCN, and KITTI is included in the SEA-RAFT TSKH mixture. KITTI fine-tuning does not help on the drone region for most supervised families, and its disadvantage grows with resolution: at $1920\times1080$ the Sintel checkpoint of RAFT, FlowFormer++, GMFlow, UniMatch and VideoFlow has a $39$--$56\%$ lower drone-region EPE than the KITTI checkpoint of the same family, and two KITTI checkpoints (VideoFlow \texttt{MOF\_kitti}, $20.674$, and VCN \texttt{kitti-ft-trainval}, $29.127$) are worse than zero flow ($7.549$). KITTI is the better domain in only three families. In ARFlow, whose KITTI checkpoints are self-supervised on raw KITTI video, every KITTI variant beats every Sintel variant at every resolution, and \texttt{KITTI12/pwclite\_ar\_mv} is the best model overall at $480\times270$ ($0.873$). In FastFlowNet and WAFT the KITTI fine-tune beats the Sintel one at every resolution; it is FastFlowNet's best checkpoint at $480\times270$ and $1920\times1080$, but never WAFT's best, where Spring-540p or the \texttt{tar-c-t} base lead. KITTI has essentially no aerial scenes, no atmospheric effects, and a strongly biased camera trajectory, and on this benchmark it does not translate into lower drone-region error for supervised fine-tuning.

\subsubsection{HD1K~\cite{kondermann2016}}
1{,}083 frames at $2560\times 1080$ of urban driving with adverse-weather variants. HD1K is a niche pre-training dataset: it never appears as a standalone fine-tune in our benchmark and is used only inside mixed training schedules, such as the \texttt{H} in SEA-RAFT's \texttt{TSKH} curriculum and as one of the data sources of UniMatch's \texttt{mixdata} variants. Because no checkpoint isolates HD1K, we cannot directly attribute performance differences to it; its effect can only be inferred jointly with Sintel and KITTI inside those mixes. UniMatch's \texttt{regrefine6-mixdata} checkpoint is the family best at $480\times270$ ($0.952$, third overall) and SEA-RAFT's TSKH releases reach the leading cluster at $1920\times1080$ ($1.883$, fourth overall), which suggests that adding HD1K's high-resolution real-world frames does no harm, but the dataset supplies the same automotive bias as KITTI and offers no aerial content.

\subsubsection{TartanAir~\cite{tartanair2020iros}}
A large-scale synthetic dataset (over $1$M frames at $640\times 480$) rendered in $30$ photorealistic Unreal Engine environments along simulated drone trajectories with aggressive 6-DoF motion (roll, yaw, pitch), day/night cycles, weather, and seasonal variation. TartanAir is used by only two of the 16 families we evaluate, it is the defining stage of SEA-RAFT (\texttt{Tartan-C-T-TSKH}) and the central refinement series in WAFT (\texttt{tar}, \texttt{tar-c}, \texttt{tar-c-t}, plus all final WAFT fine-tunes). No other model in the benchmark has been exposed to TartanAir. The two families behave differently with resolution. SEA-RAFT climbs steadily, from tenth at $480\times270$ ($1.117$) to fifth at $960\times432$ ($1.557$) and fourth at $1920\times1080$ ($1.883$), while WAFT is mid-table throughout (sixth, ninth and eighth; $1.000$, $1.793$, $2.086$). Both belong to the leading cluster at $1920\times1080$, but neither leads at any resolution: the first ranks go to ARFlow and UniMatch, which have never seen TartanAir. Aerial pre-training is therefore compatible with strong high-resolution performance but is not what decides the top of the leaderboard.

\subsubsection{Spring~\cite{mehl2023}}
A high-resolution ($1920\times1080$) cinematic synthetic dataset rendered in Blender, with stereo + forward/backward optical flow. It complements Sintel with sharper detail and higher-resolution displacements but inherits Sintel's cinematic, character-animation bias. Spring has very limited adoption in the families we evaluate: only WAFT (\texttt{tar-c-t-spring-540p}, \texttt{tar-c-t-spring-1080p}) and MemFlow (\texttt{MemFlowNet\_spring}) include a Spring fine-tune (SEA-RAFT additionally releases a Spring variant of its TSKH curriculum). In both families the Spring fine-tune is the best checkpoint at the two lower resolutions: WAFT \texttt{tar-c-t-spring-540p} reaches $1.000$ and $1.793$, and \texttt{MemFlowNet\_spring} reaches $0.948$ (second overall at $480\times270$) and $1.484$. At $1920\times1080$ the Spring advantage disappears: WAFT's best checkpoint is the \texttt{tar-c-t} base without a final fine-tune ($2.086$ versus $2.428$ for Spring-540p), and \texttt{MemFlowNet\_spring} ($2.016$) falls just behind \texttt{MemFlowNet\_T\_sintel} ($1.965$). Training at Spring's native resolution does not pay off either: \texttt{tar-c-t-spring-1080p} is worse than \texttt{tar-c-t-spring-540p} at all three resolutions ($1.099$/$1.830$/$2.734$ versus $1.000$/$1.793$/$2.428$).

\subsubsection{CityScapes~\cite{cordts2016}}
Real driving images at $2048\times 1024$ with no dense optical-flow ground truth. CityScapes is used by exactly one family in the benchmark, ARFlow, and only for self-supervised consistency training (\texttt{CityScapes/pwclite\_ar}); no other model in our study has been exposed to it. The training signal therefore reflects driving-scene content similar to KITTI's. ARFlow's CityScapes checkpoint sits between its siblings at every resolution: behind every KITTI-trained ARFlow checkpoint and ahead of every Sintel-trained one ($0.940$, $1.748$, $3.442$ against $0.873$/$1.581$/$3.175$ for the best KITTI and $0.996$/$1.916$/$3.756$ for the best Sintel variant). Within ARFlow's self-supervised objective, driving-scene data is thus consistently preferable to Sintel, the reverse of the pattern in most supervised families.

\subsubsection{Implications for FlyingDrones}
Three observations summarise the schedule effects we observe.
First, for supervised fine-tuning, Sintel is the most reliable last stage, and its advantage over KITTI grows with resolution (median $3.5\%$, $20.0\%$ and $22.0\%$ lower drone-region EPE at the three resolutions). This is not a matter of displacement range: KITTI's labelled displacements are larger than Sintel's (median $23$ vs.\ $3.9$\,px) and than the drone motion in our benchmark, so the Sintel advantage comes from properties other than the range of motion; KITTI is preferable only for the self-supervised ARFlow and for FastFlowNet and WAFT.
Second, no public pre-training dataset combines (i) genuine aerial ego-motion, (ii) atmospheric effects and natural lighting, and (iii) a small-displacement-dominated tail at high resolution. TartanAir comes closest on (i) but is rendered at $480$\,px and is not available to most curricula; ChairsSDHom matches (iii) but lacks (i) and (ii); Sintel and Spring match (ii) but not (iii). The $1920\times1080$ regime in our benchmark therefore has no closely matched fine-tuning dataset: the best checkpoint there (UniMatch \texttt{regrefine6-sintelft}) removes $79\%$ of the zero-flow error but still leaves $1.557$\,px on the drone region, and the families outside the leading cluster leave $3.1$\,px or more. Appendix~\ref{sec:finetune} tests directly whether in-domain FlyingDrones fine-tuning closes part of this gap.
Third, longer temporal windows do not compensate for the missing domain-matched data: VideoFlow's five-frame MOF checkpoints improve on their three-frame BOF counterparts in only one of twelve matched comparisons, and at $1920\times1080$ the three-frame \texttt{BOF\_sintel} is statistically tied with the two-frame UniMatch \texttt{regrefine6-sintelft}. On this benchmark, the final fine-tuning dataset and the matching architecture matter more than the number of input frames.

\subsection{Training Curricula and Their Effect on Drone-Region Error}
\label{sec:appendix-curricula-discussion}

The 16 model families in this benchmark span four broad training paradigms: classical pyramid-style supervised flow, recurrent correlation refinement, transformer-style global matching, and temporally-extended or generative formulations. The aggregated full-frame and drone-region tables (Tables~\ref{tab:agg-full}--\ref{tab:agg-drone}) make clear that, on tiny aerial targets, the choice of checkpoint within a family often matters as much as the architecture. Read jointly with the dataset summary in Appendix~\ref{sec:appendix-datasets}, the picture is: for supervised families a Sintel fine-tune is the most reliable last stage and its advantage over KITTI grows with resolution; Things-stage checkpoints remain competitive; TartanAir pre-training is compatible with, but not decisive for, high-resolution performance; and no public dataset matches the $1920\times1080$ aerial regime. In what follows we walk through the families in the same order as the appendix tables, tying each release naming convention to its observed behaviour and to the datasets that drove it.

\subsubsection{Classical and pyramid-style baselines}
PyFlow's \texttt{accurate}, \texttt{default}, and \texttt{fast} entries are runtime presets of the same Ce~Liu coarse-to-fine variational solver, with no learned weights. They isolate what a smoothness prior alone achieves: at $480\times270$, where most drone motion is sub-pixel, \texttt{fast} is fourth overall ($0.977$), but its confidence interval overlaps those of most of the field (ranks 1--14), and as the drone's displacement grows the solver falls to tenth at $960\times432$ ($1.895$) and thirteenth at $1920\times1080$ ($4.051$). The faster presets are better on the drone region at every resolution ($0.977$ for \texttt{fast} versus $1.175$ for \texttt{accurate} at $480\times270$, $4.051$ versus $4.520$ at $1920\times1080$). FlowNet2 follows the well-known \texttt{C}\,$\rightarrow$\,\texttt{CS}\,$\rightarrow$\,\texttt{CSS-ft-sd} curriculum, with \texttt{S}/\texttt{SD} as released single-stream variants; the single-stream \texttt{S} is the weakest checkpoint at every resolution and the Chairs-only \texttt{C} is among the three weakest, while the full \texttt{FlowNet2} ensemble ($1.209$, $3.136$) and \texttt{CS} ($2.011$ at $960\times432$) lead the family. PWC-Net is the shortest curriculum we evaluate, only \texttt{pwc\_net\_chairs} and the published \texttt{pwc\_net} refinement. The Chairs checkpoint is better on the drone region at the two lower resolutions ($1.227$ vs.\ $1.356$, $2.125$ vs.\ $2.546$) but collapses at $1920\times1080$ ($9.768$, above the zero-flow $7.549$), where the refinement brings it back to $4.404$ (full frame $3.750\rightarrow1.215$). LiteFlowNet3 splits its release into a smaller (\texttt{3s}) and a larger (\texttt{3}) network, each with a Sintel and a KITTI fine-tune. The larger Sintel model is the family best at every resolution ($1.192$, $2.147$, $3.275$), the Sintel fine-tune beats the KITTI one for both network sizes at every resolution, and the KITTI gap widens with resolution ($3.751$ for \texttt{liteflownet3-kitti} at $1920\times1080$). Network size changes drone-region EPE by at most $0.21$\,px, less than the choice of fine-tune. VCN's release paths separate early \texttt{chairs} and \texttt{things} models from \texttt{sintel-ft-trainval} and \texttt{kitti-ft-trainval} fine-tunes. At $480\times270$ all seven checkpoints lie within $1.067$--$1.246$ (\texttt{things/211999} best, seventh overall), and at $960\times432$ the Sintel fine-tune is best ($2.048$) while \texttt{kitti-ft} is already worse than zero flow ($4.792$ vs.\ $3.498$). At $1920\times1080$, where VCN runs on the untiled full frame, every Things or fine-tuned checkpoint breaks down ($13.459$--$31.172$, all above zero flow) and only the two \texttt{chairs} checkpoints remain usable ($3.829$ and $4.143$), giving the family an $8.1\times$ spread. Across these families, the pyramidal smoothness assumption is competitive while drone motion is sub-pixel but does not scale: at $1920\times1080$ all of them sit at $3.1$\,px or above, against $\leq 2.09$\,px for the leading cluster.

\subsubsection{Self-supervised and lightweight variants}
ARFlow exposes the training domain directly through its checkpoint names: \texttt{CityScapes/...}, \texttt{KITTI12/...}, \texttt{KITTI15/...}, and \texttt{Sintel/...}, with \texttt{pwclite\_raw} the raw self-supervised model, \texttt{pwclite\_ar} the analogy-augmented two-frame model, and \texttt{pwclite\_ar\_mv} the three-frame multi-view extension. Because the supervision is unsupervised on the named domain, each entry is essentially a head-to-head comparison of target-domain adaptation. Every KITTI checkpoint beats every Sintel checkpoint at every resolution, with CityScapes in between, and the family spread is narrow ($1.21$--$1.29\times$). \texttt{KITTI12/pwclite\_ar\_mv} is the best model overall at $480\times270$ ($0.873$, ahead of MemFlow by $0.075$\,px in every bootstrap resample) and has the lowest full-frame EPE of any checkpoint at all three resolutions ($0.150$, $0.304$, $0.684$). Its drone-region rank, however, falls to sixth at $960\times432$ ($1.581$) and to tenth at $1920\times1080$ (best ARFlow checkpoint \texttt{KITTI15/pwclite\_ar}, $3.175$), so the full-frame advantage does not reflect more accurate motion on the target; Section~\ref{sec:efficient} attributes it to near-zero output on static background. The multi-view extension changes drone-region EPE by at most $0.22$\,px and is not uniformly better (it helps for KITTI12 and Sintel, but not for KITTI15 at the two higher resolutions). FastFlowNet uses the conventional \texttt{chairs}\,$\rightarrow$\,\texttt{things3d} synthetic curriculum and adds three fine-tunes (\texttt{ft\_sintel}, \texttt{ft\_kitti}, \texttt{ft\_mix}). \texttt{ft\_kitti} is the family best at $480\times270$ and $1920\times1080$ ($1.459$, $4.425$) and \texttt{things3d} at $960\times432$ ($2.336$), while \texttt{things3d} and \texttt{chairs} collapse at $1920\times1080$ ($8.333$ and $7.432$, against $7.549$ for zero flow). With these best checkpoints FastFlowNet is still last at the two lower resolutions and fifteenth at $1920\times1080$, so in this regime the lightweight pyramid does cost accuracy.

\subsubsection{Correlation-and-refinement family}
RAFT is the cleanest example of a staged curriculum: \texttt{raft-chairs}, \texttt{raft-things}, \texttt{raft-sintel}, \texttt{raft-kitti} expose the entire FlyingChairs $\rightarrow$ FlyingThings3D $\rightarrow$ Sintel $\rightarrow$ KITTI sequence, while \texttt{raft-small} retains the earlier synthetic stages in a reduced-capacity model. At the two higher resolutions drone-region EPE falls monotonically through the curriculum up to Sintel and rises again with the final KITTI stage ($2.149\rightarrow1.811\rightarrow1.713\rightarrow2.352$ at $960\times432$; $3.280\rightarrow2.418\rightarrow2.047\rightarrow3.378$ at $1920\times1080$); at $480\times270$ \texttt{raft-things} and \texttt{raft-sintel} are tied ($1.152$, $1.160$). The family's spread grows from $1.2\times$ to $1.9\times$ with resolution, and \texttt{raft-small} is the weakest checkpoint at every resolution, its gap to \texttt{raft-things} widening from $0.25$ to $1.42$\,px. With its best checkpoint RAFT ranks eleventh, eighth and seventh. SEA-RAFT moves the curriculum toward navigation: its releases stack TartanAir, FlyingChairs, FlyingThings3D, and a TSKH (Things+Sintel+KITTI+HD1K) mixed refinement, and are published as three distinct KITTI, Sintel and Spring checkpoints. The Sintel release is the family best at every resolution ($1.117$, $1.557$, $1.883$), ahead of the KITTI ($1.160$, $1.786$, $2.026$) and Spring ($1.189$, $2.003$, $2.972$) releases, and the family climbs steadily with resolution (tenth, fifth, fourth), reaching the leading cluster at $1920\times1080$ without tiled inference. WAFT layers a frozen DepthAnythingV2~\cite{depth_anything_v2} encoder on a Chairs+Things base and then progressively refines through \texttt{tar} (TartanAir), \texttt{tar-c}, and \texttt{tar-c-t}, with optional final fine-tunes on Sintel, KITTI, or Spring at 540p or 1080p. The best checkpoint is \texttt{tar-c-t-spring-540p} at the two lower resolutions ($1.000$, $1.793$) and the \texttt{tar-c-t} base at $1920\times1080$ ($2.086$), placing WAFT sixth, ninth and eighth. The Sintel fine-tune is the family's worst at every resolution ($1.493$, $2.383$, $4.657$; a $2.2\times$ spread at $1920\times1080$), and its full-frame EPE at the two higher resolutions ($6.449$, $5.636$) is even worse than zero flow ($0.973$, $2.065$), the clearest case in the benchmark of a harmful Sintel fine-tune.

\subsubsection{Transformer-style and global-matching models}
GMFlow's release set spans both base checkpoints (\texttt{...chairs...}, \texttt{...things...}) and benchmark-specialized fine-tunes (\texttt{...sintel...}, \texttt{...kitti...}), each with and without the \texttt{with\_refine} head, so it is well suited to isolating the effect of the last training stage. For the Chairs, Things and Sintel checkpoints it lowers drone-region EPE at every resolution, by $1.10$--$1.35$\,px at $1920\times1080$, whereas for the KITTI checkpoints it makes no consistent difference. The family best is \texttt{with\_refine\_things} at $480\times270$ ($1.282$, fifteenth overall) and \texttt{with\_refine\_sintel} at the two higher resolutions ($1.665$, $1.996$; seventh and sixth), while the KITTI fine-tunes are the weakest refined checkpoints at the two higher resolutions ($2.694$, $3.575$). Three GMFlow checkpoints, including \texttt{with\_refine\_things} and \texttt{with\_refine\_sintel}, are the same released weights as UniMatch's \texttt{gmflow-scale1-things}, \texttt{gmflow-scale2-things} and \texttt{gmflow-scale2-sintel} and agree with them up to floating-point noise. UniMatch shares the GMFlow backbone and extends it with a second matching scale, six regression-refinement steps (\texttt{regrefine6}), and \texttt{mixdata} (multi-dataset) checkpoints between the synthetic Things stage and the final \texttt{sintelft}/\texttt{kitti15} fine-tunes. Each architectural addition lowers drone-region EPE at every resolution for the same training data: on Things, \texttt{scale1} $\rightarrow$ \texttt{scale2} $\rightarrow$ \texttt{scale2-regrefine6} goes from $3.545$ to $2.199$ to $1.658$ at $1920\times1080$. The \texttt{regrefine6} checkpoints are the family best at every resolution, with the leading training set changing with resolution: \texttt{mixdata} at $480\times270$ ($0.952$, third overall), Things at $960\times432$ ($1.293$, first, ahead of VideoFlow in $>99\%$ of resamples), and \texttt{sintelft} at $1920\times1080$ ($1.557$, first, but tied with VideoFlow \texttt{BOF\_sintel}). The \texttt{kitti15} fine-tune of the same architecture reaches only $3.105$ at $1920\times1080$, twice the error of \texttt{sintelft}. FlowFormer++ keeps a similar synthetic-then-benchmark schedule, with \texttt{things} and \texttt{things\_288960} (a Things-stage release at the $288\times960$ KITTI-style crop) preceding the \texttt{sintel} and \texttt{kitti} fine-tunes. The Things releases are the family best at every resolution ($1.108$, $1.513$, $1.840$; ninth, fourth and third overall), Sintel follows ($1.201$, $1.880$, $2.002$), and KITTI is clearly worst ($1.364$, $2.782$, $3.487$), giving a $1.8\times$ and $1.9\times$ within-architecture swing at the two higher resolutions.

\subsubsection{Temporal, memory-based, and generative models}
VideoFlow's checkpoints fall into two families distinguished by temporal window: \texttt{BOF} (three-frame) and \texttt{MOF} (five-frame). Each is released in synthetic Things and in Sintel- and KITTI-finetuned variants. Three findings are robust. First, the longer window does not help: the BOF checkpoint is better than its MOF counterpart in 11 of the 12 matched checkpoint\,$\times$\,resolution comparisons, the exception being \texttt{MOF\_things\_288960noise} at $480\times270$ ($0.997$ vs.\ $1.056$, the family's leaderboard entry there). Second, \texttt{BOF\_sintel} is the family best at the two higher resolutions ($1.425$, $1.582$; second overall at both, tied with UniMatch at $1920\times1080$), with the Things variants close behind. Third, the KITTI fine-tunes are the weakest, and \texttt{MOF\_kitti} collapses beyond the zero-flow level at $960\times432$ ($4.210$) and $1920\times1080$ ($20.674$), so VideoFlow is simultaneously one of the benchmark's best families and its most checkpoint-sensitive one ($13\times$ spread at $1920\times1080$). MemFlow shares the Things\,$\rightarrow$\,\{Sintel, KITTI, Spring\} curriculum but adds an explicit memory module; its T-prefix variants train the memory across an additional temporal stage. \texttt{MemFlowNet\_spring} is the family best at the two lower resolutions ($0.948$, second overall; $1.484$, third) and \texttt{MemFlowNet\_T\_sintel} at $1920\times1080$ ($1.965$, fifth). The T-prefix variants are not consistently better than their plain counterparts, and \texttt{MemFlowNet\_T\_kitti} breaks down at $1920\times1080$ ($6.778$, with a full-frame EPE of $7.444$ against $2.065$ for zero flow), giving MemFlow a $3.5\times$ spread there. MemFlow thus stays in the leading group at every resolution, but its memory does not lift it above the best two- and three-frame models at the higher resolutions. FlowDiffusion (DDVM) is released only as two pipeline snapshots, \texttt{pipeline-305000} and \texttt{pipeline-900000}, which differ in optimization steps rather than in named benchmark curricula. Their drone-region EPE values differ by less than $0.05$\,px at every resolution ($1.108$/$1.099$, $2.220$/$2.228$, $5.125$/$5.173$), so additional denoising training yields no drone-region gain; the family is eighth at $480\times270$ but falls to fifteenth and last at the two higher resolutions.

\subsubsection{Summary across curricula}
Four patterns emerge across all 16 families. First, the choice of checkpoint within a family moves drone-region error as much as the choice of architecture: at $960\times432$ the within-family spreads of VideoFlow ($3.0\times$), VCN ($2.3\times$) and UniMatch ($2.1\times$) exceed the $1.8\times$ span of the entire best-per-family leaderboard, and at $1920\times1080$ VideoFlow ($13\times$, from \texttt{MOF\_kitti}) and VCN ($8.1\times$, untiled) far exceed the leaderboard's $3.3\times$. Second, for supervised families Sintel is the better of the two near-universal fine-tunes, in 8, 9 and 9 of the 12 families that release both, with a median advantage that grows from $3.5\%$ to $20.0\%$ and $22.0\%$; KITTI wins consistently only for ARFlow, FastFlowNet and WAFT. Third, neither longer temporal windows nor aerial pre-training decide the top of the leaderboard: VideoFlow MOF does not improve on BOF, the two-frame UniMatch leads at $960\times432$ and shares the lead at $1920\times1080$, and the TartanAir-trained SEA-RAFT and WAFT reach the leading cluster at $1920\times1080$ without leading at any resolution. Fourth, optimization-stage snapshots (FlowDiffusion, $<0.05$\,px), ARFlow's multi-view extension ($\leq0.22$\,px) and LiteFlowNet3's network size ($\leq0.21$\,px) produce much smaller drone-region differences than the final fine-tuning dataset, whereas architectural refinements within one release (GMFlow's refinement head, UniMatch's second scale and \texttt{regrefine6}) and reduced capacity (\texttt{raft-small}) matter more as resolution grows. Appendix~\ref{sec:real} examines how these curriculum effects carry over to real footage.

\subsection{Sensitivity to the Sampling Protocol}
\label{sec:appendix-protocol}

Because the stratified subset described in Section~\ref{sec:results-protocol-anchor} is a design choice rather than the only reasonable way to sample the sequence, we check its effect on the reported rankings against the simplest alternative: a contiguous window of the first 500 evaluable reference frames (frames 1--500, each paired with its predecessor). Table~\ref{tab:protocol} compares the two samples on the leaderboard checkpoint of each family, holding the checkpoint fixed so that any difference is attributable to the sampling protocol alone rather than to checkpoint selection. The window is harder for every family at every resolution: drone-region EPE is higher by a median of $0.69$, $1.18$ and $1.29$\,px (mean $0.69$, $1.15$ and $1.64$\,px) at $480\times270$, $960\times432$ and $1920\times1080$. The zero-flow reference explains the direction of this shift: its drone-region EPE, i.e.\ the mean ground-truth drone motion, is $25$--$28\%$ larger on the window ($2.229$, $4.356$, $9.655$\,px versus $1.744$, $3.498$, $7.549$\,px), so the opening frames over-represent large drone displacements relative to the stratified design. At $1920\times1080$ the shift is smallest for the leading families (UniMatch, VideoFlow, MemFlow and GMFlow move by $0.22$--$0.37$\,px) and largest for those that already struggle at that resolution (up to $4.06$\,px for FlowDiffusion).

The family ranking is largely preserved: the Spearman rank correlation between the two samples is $0.83$, $0.93$ and $0.95$ at the three resolutions. At $960\times432$ and $1920\times1080$ the same two families, UniMatch and VideoFlow, lead under both protocols (their order swaps at $960\times432$). At $480\times270$, where ranks two to six of the stratified leaderboard lie within about $0.05$\,px of each other, the order at the top changes (PyFlow, UniMatch and ARFlow lead the window), consistent with the overlapping confidence intervals of the stratified leaderboard. Because the best checkpoint per family can differ between samples, all leaderboard tables select the best checkpoint on the stratified subset used throughout the paper.

\input{tables/tab_protocol}

\subsection{Real-Footage Ranking Transfer}\label{sec:real}

The benchmark above isolates optical-flow quality from downstream detection and tracking, and, because real drone footage lacks dense ground-truth flow, it does so on synthetic imagery. We complement it with a ground-truth-free study on real UAV footage that tests whether its rankings and findings survive contact with real video.

We use the multi-view drone-tracking dataset of Li et al.~\cite{li_reconstruction_2020}, in which a hexacopter is filmed by static, tripod-mounted consumer cameras with a manually labelled 2-D image position of the drone centre in every frame. Because the camera is static, the drone is essentially the only moving object, so a faithful flow field must place a localised motion response on it. We evaluate on all 21 labelled cameras of datasets 1--4 (dataset 5 has no 2-D labels), drawing 100 frame pairs per camera stratified by the drone's image-space displacement $\lVert\mathbf{d}\rVert$ (the label motion, valid as pseudo-ground-truth because the camera is static) into sub-pixel, low, mid, and high bins. Each checkpoint is scored per camera and we report mean~$\pm$~std across the 21 cameras, so the spread is a genuine scene/viewpoint error bar. Treating the flow-magnitude field as a motion-saliency map and the label as the target, we report NSS (normalized scanpath saliency at the drone, i.e.\ how many standard deviations the drone region stands out as a motion outlier; the drone-disk variant $\mathrm{NSS}_D$ tolerates label noise), Hit@$r$ (fraction of frames whose smoothed flow-magnitude peak lies within $r$\,px of the drone), contrast (mean flow magnitude on the drone versus background), and pt-EPE/pt-cos (endpoint error and cosine similarity between the predicted flow at the drone and the label displacement, an absolute check of motion fidelity). We evaluate 41 published checkpoints spanning all 16 families at $1280\times720$ (VideoFlow's five-frame MOF at $960\times540$ for memory), rescaling every field to native $1920\times1080$; the complete per-checkpoint table is in Appendix~\ref{app:real} (Table~\ref{tab:real-all}).

The synthetic and real orderings agree moderately, and checkpoint choice again matters as much as architecture. Across the 41 checkpoints, real $\mathrm{NSS}_D$ and synthetic drone-region EPE are positively rank-correlated at every synthetic resolution (Spearman $\rho=0.54$, $0.63$ and $0.60$ against $480\times270$, $960\times432$ and $1920\times1080$; all $p<10^{-3}$), most strongly at $960\times432$, the synthetic resolution closest to the real inference scale. Agreement is weaker at family level ($\rho=0.38$--$0.51$ over the 16 best-per-family entries), partly because several families' strongest synthetic checkpoints were not part of the real study. MemFlow is the clearest case of agreement: \texttt{spring} is the strongest real system ($\mathrm{NSS}_D=27.3\pm10.2$, Hit@50 $=0.62$), all four MemFlow checkpoints are in the real top eight, and MemFlow is in the synthetic top five at every resolution. A classical variational method is nearly as strong on real footage: PyFlow ranks second ($26.1\pm9.9$) and attains the highest Hit@50 ($0.71$), although it is only mid-field on synthetic data at the two higher resolutions; having no learned prior, it does not hallucinate motion on the low-contrast sky. The largest disagreement is VideoFlow: its only real-evaluated checkpoint, the five-frame \texttt{MOF\_sintel}, is in the lower third ($4.9\pm4.4$, Hit@50 $=0.16$), whereas the family's three-frame \texttt{BOF\_sintel} is second on synthetic data at the two higher resolutions; fine-tuning on FlyingDrones lifts the same MOF model to $17.8$ (Appendix~\ref{sec:finetune}). Checkpoint choice within one architecture swings real $\mathrm{NSS}_D$ by up to $2.8\times$ (FlowFormer++, $15.5$ for \texttt{things\_288960} vs.\ $5.5$ for \texttt{kitti}) and by about $2\times$ for GMFlow, UniMatch, LiteFlowNet3, FastFlowNet, MemFlow and RAFT.

RAFT transfers across all four checkpoints ($\mathrm{NSS}_D\in[8.6,15.5]$, Hit@50 up to $0.52$, pt-EPE $1.6$--$2.2$\,px), consistent with its mid-to-upper synthetic standing, and SEA-RAFT is third on real footage ($16.5$). Sintel-trained checkpoints are not systematically worse on real footage: RAFT \texttt{sintel}, MemFlow \texttt{sintel} and SEA-RAFT are all in the top five, and FlowFormer++ \texttt{sintel} reaches $12.0$. The weakest transfer comes from early stacked networks and the diffusion model (FlowNet2-S last at $0.96$, FlowNet2-C $1.54$, FlowDiffusion $2.93$). Localisation is gated by motion magnitude, as the synthetic sub-pixel analysis predicts: for the transferable checkpoints, Hit@50 rises from weak in the sub-pixel regime to near-perfect once the drone moves more than a few pixels (e.g.\ MemFlow \texttt{spring} $0.24\to0.90$, PyFlow $0.54\to0.88$, RAFT \texttt{sintel} $0.08\to0.89$), while VideoFlow \texttt{MOF\_sintel} stays low at every level ($\leq0.39$), a qualitatively different failure (full breakdown in Appendix~\ref{app:real}, Table~\ref{tab:real-strata}).

Separating what transfers from what does not: the regime-level claims hold (sub-pixel motion is the binding constraint, drone-region evaluation is necessary), the synthetic ranking carries a moderate but significant signal about real-world motion localisation, and checkpoint choice remains a first-order factor in both domains. What does not transfer reliably is the exact top of the ranking: classical PyFlow is far stronger on real footage than on synthetic data at the higher resolutions, and the five-frame VideoFlow variant transfers poorly without in-domain training. A synthetic benchmark of this kind is therefore best used to shortlist architectures and curricula, followed by validation on in-domain footage before a checkpoint is deployed; the motion-alignment proxy used here is itself a cheap, label-only tool for that validation. Resolution-matched inference, so that the drone retains supra-pixel motion, and treating the sub-pixel regime as a hard floor where detection must fall back on appearance or temporal accumulation follow from both studies.

\subsection{Real-Footage Motion-Alignment Results}
\label{app:real}

Table~\ref{tab:real-all} reports the motion-alignment metrics described above for every published checkpoint, on 100 displacement-stratified frame pairs per camera from the static-camera real UAV sequences of~\cite{li_reconstruction_2020} ($1280\times720$ inference, VideoFlow MOF at $960\times540$, all fields rescaled to $1920\times1080$). $\mathrm{NSS}_D$ is the drone-disk normalized scanpath saliency (higher is better); $\mathrm{Hit}@50$ is the fraction of frames whose border-masked flow-magnitude peak falls within $50$\,px of the labelled drone centre; \emph{Contr.} is the drone/background magnitude ratio; \emph{cos} and \emph{pt-EPE} are the direction agreement and endpoint error against the label displacement. Rows are ordered by $\mathrm{NSS}_D$; every system is evaluated on all 21 cameras, and every checkpoint is loaded with a check that all network weights match the released file. UniMatch \texttt{scale1-things} and GMFlow \texttt{things} share identical weights and hence identical scores.

\input{tables/tab_real_all}

Table~\ref{tab:real-strata} breaks down localisation by the drone's image-space displacement stratum for representative checkpoints, referenced above.

\input{tables/tab_real_strata}

\subsection{Fine-Tuning on FlyingDrones}\label{sec:finetune}

The benchmark so far measures existing checkpoints. As a next step, we fine-tune four representative architectures on a held-out split of FlyingDrones and evaluate them on synthetic data never seen during training and on the real footage of Section~\ref{sec:real}. This experiment additionally serves as a controlled comparison with all four models are starting from a public initialisation, see the identical training pairs, and are optimised under the same schedule.

\subsubsection{Setup}

We partition the 1000-frame stratified evaluation subset into 800 train / 100 validation / 100 test frames by a second, nested stratification that keeps every stratum represented in all three parts; the split is released with the dataset. Because neighbouring frames of the sequence are nearly identical, frames that lie four or fewer frames apart are first chained into blocks and whole blocks are assigned to one partition, so every test and validation frame is at least five frames away from any training frame and no five-frame window crosses the partition. Each frame $t$ supervises the pair $(I_t, I_{t-1})$ with its backward ground truth (799 training pairs, as frame $0$ has no predecessor); VideoFlow MOF is trained on the genuine time-reversed five-frame window $[I_{t+2},\dots,I_{t-2}]$ with ground truth on the three forward slots. The models are initialised from FlowFormer++ \texttt{things\_288960}, GMFlow \texttt{gmflow\_chairs}, RAFT \texttt{raft-things} and VideoFlow \texttt{MOF\_sintel}. All four are trained for 1500 AdamW steps with a one-cycle schedule and mixed precision on random crops from frames resized to the evaluation scale; per-arm batch, crop, learning rate and loss weighting are listed in Table~\ref{tab:ft-curves}. Validation runs every 150 steps using the evaluation path (the same code that produces the benchmark tables) and the reported model is the one minimising validation drone-region EPE. Test scores are computed on the 100 held-out pairs with that same canonical evaluator, and the same two checkpoints per arm are evaluated on real footage with the protocol of Section~\ref{sec:real}.

We select four architectures spanning the design space: FlowFormer++ (transformer decoding), GMFlow (global matching, fine-tuned at $1280\times720$), RAFT (correlation-based recurrent refinement) and VideoFlow MOF (a five-frame temporal model).

\subsubsection{Results}

Table~\ref{tab:ft-curves} reports the recipe and optimisation behaviour and Table~\ref{tab:ft-test} the held-out synthetic and real scores.

Three of the four architectures gain substantially on held-out synthetic data. VideoFlow MOF improves drone-region EPE from $1.727$ to $1.089$ ($-36.9\%$), GMFlow from $3.649$ to $2.723$ ($-25.4\%$, at $1280\times720$) and FlowFormer++ from $1.614$ to $1.305$ ($-19.2\%$); full-frame EPE falls by $88\%$ for VideoFlow and FlowFormer++ and by $43\%$ for GMFlow. Eight hundred in-domain pairs (less than $9\%$ of the dataset) are therefore sufficient to substantially reduce the drone-region error of a strong public checkpoint, direct evidence that FlyingDrones carries a learnable, task-specific signal rather than merely re-ranking existing models. RAFT is the exception: its validation drone-region EPE never improves on the public \texttt{raft-things} initialisation, the selected checkpoint is the first one (step 150), and the held-out drone-region EPE is marginally worse ($1.857\rightarrow1.933$, $+4.1\%$) even though its full-frame EPE improves by $16\%$.

The gains transfer to real footage. All four fine-tuned checkpoints localise the drone better than their initialisation on the 21 real cameras: $\mathrm{NSS}_D$ rises from $4.9$ to $17.8$ for VideoFlow MOF (Hit@50 $0.16\rightarrow0.56$), from $15.5$ to $22.6$ for FlowFormer++, from $4.4$ to $6.8$ for GMFlow and from $14.5$ to $16.9$ for RAFT. Inserted into the ranking of the 41 public checkpoints, fine-tuned FlowFormer++ and VideoFlow MOF would both rank third. For VideoFlow the improvement is visible in every displacement stratum (Hit@50 $0.07\rightarrow0.21$ sub-pixel, $0.11\rightarrow0.58$ low, $0.25\rightarrow0.79$ mid, $0.39\rightarrow0.80$ high).

Taken together, the study shows that fine-tuning on FlyingDrones improves held-out synthetic accuracy for three of four architectures and real-footage motion localisation for all four, with the largest real gain for the temporal model that transferred worst out of the box. The benefit is not uniform across architectures, mirroring the regime-dependence the benchmark itself exposes, and the real-footage evaluation remains a static-camera proxy (see Limitations).

\input{tables/tab_ft}

\pagebreak
\FloatBarrier

\input{checklist}

\end{document}

%% file: tables/tab_leaderboard_main.tex
\begin{table}[t]
\centering
\caption{Best drone-region EPE (px) per model family at each resolution on the stratified subset (999 flow pairs; lower is better). The best checkpoint is selected per family and resolution, so it can differ across columns. Families are sorted by their mean rank $\bar r$ over the three resolutions. \emph{Cls}: C~= correlation/iterative, P~= classical\,/\,pyramid, E~= efficient\,/\,reduced-supervision, T~= transformer\,/\,global matching, M~= temporal\,/\,memory\,/\,generative. $\dagger$~more than two input frames; $\ddagger$~tiled inference ($960\times432$ tiles; $960\times288$ for FlowFormer++ \texttt{things\_288960}). Selected checkpoints, 95\% bootstrap confidence intervals and further metrics are in Tables~\ref{tab:summary-best-drone-270x480}--\ref{tab:summary-best-drone-1080x1920}.}
\label{tab:leaderboard-main}
\small
\setlength{\tabcolsep}{6pt}
\begin{tabular}{clrrrr}
\toprule
Cls & Family & 480$\times$270 & 960$\times$432 & 1920$\times$1080 & $\bar r$ \\
\midrule
T & UniMatch & 0.952 & 1.293 & 1.557$^\ddagger$ & 1.7 \\
M & VideoFlow$^\dagger$ & 0.997 & 1.425 & 1.582$^\ddagger$ & 3.0 \\
M & MemFlow$^\dagger$ & 0.948 & 1.484 & 1.965$^\ddagger$ & 3.3 \\
T & FlowFormer++ & 1.108 & 1.513 & 1.840$^\ddagger$ & 5.3 \\
E & ARFlow & 0.873 & 1.581 & 3.175\phantom{$^\ddagger$} & 5.7 \\
C & SEA-RAFT & 1.117 & 1.557 & 1.883\phantom{$^\ddagger$} & 6.3 \\
C & WAFT & 1.000 & 1.793 & 2.086$^\ddagger$ & 7.7 \\
C & RAFT & 1.152 & 1.713 & 2.047$^\ddagger$ & 8.7 \\
P & PyFlow (C2F) & 0.977 & 1.895 & 4.051\phantom{$^\ddagger$} & 9.0 \\
T & GMFlow & 1.282 & 1.665 & 1.996$^\ddagger$ & 9.3 \\
P & VCN & 1.067 & 2.048 & 3.829\phantom{$^\ddagger$} & 10.3 \\
P & FlowNet2 & 1.209 & 2.011 & 3.136\phantom{$^\ddagger$} & 11.0 \\
P & LiteFlowNet3 & 1.192 & 2.147 & 3.275$^\ddagger$ & 12.3 \\
M & FlowDiffusion (DDVM) & 1.099 & 2.220 & 5.125\phantom{$^\ddagger$} & 13.0 \\
P & PWC-Net & 1.227 & 2.125 & 4.404\phantom{$^\ddagger$} & 13.7 \\
E & FastFlowNet & 1.459 & 2.336 & 4.425\phantom{$^\ddagger$} & 15.7 \\
\midrule
-- & Zero flow (reference) & 1.744 & 3.498 & 7.549\phantom{$^\ddagger$} & -- \\
\bottomrule
\end{tabular}
\end{table}

%% file: tables/tab_ft_main.tex
\begin{table}[t]
\centering
\caption{Fine-tuning on 800 FlyingDrones pairs: public initialisation $\rightarrow$ fine-tuned. Synthetic: drone-region EPE on the 100-pair held-out test split (lower better). Real: motion alignment on 21 real cameras (higher better). Per-arm recipes and full-frame EPE in Tables~\ref{tab:ft-curves} and~\ref{tab:ft-test}.}
\label{tab:ft-main}
\small
\setlength{\tabcolsep}{4pt}
\begin{tabular}{lccc}
\toprule
Model & Synth.\ drone EPE ($\Delta$) & Real $\mathrm{NSS}_D$ & Real Hit@50 \\
\midrule
FlowFormer++ & $1.61\rightarrow1.30$ ($-19.2\%$) & $15.5\rightarrow22.6$ & $0.56\rightarrow0.58$ \\
GMFlow & $3.65\rightarrow2.72$ ($-25.4\%$) & $4.4\rightarrow6.8$ & $0.23\rightarrow0.31$ \\
RAFT & $1.86\rightarrow1.93$ ($+4.1\%$) & $14.5\rightarrow16.9$ & $0.52\rightarrow0.53$ \\
VideoFlow MOF & $1.73\rightarrow1.09$ ($-36.9\%$) & $4.9\rightarrow17.8$ & $0.16\rightarrow0.56$ \\
\bottomrule
\end{tabular}
\end{table}

%% file: tables/tab_summary_best_drone_270x480.tex
\begin{table}[h]
  \caption{Best drone-region EPE per model family at 480$\times$270, on the stratified subset (999 flow pairs). One row per family: the checkpoint minimising drone-region EPE.Brackets give the 95\% percentile-bootstrap confidence interval of the drone-region EPE (10{,}000 resamples over pairs). \emph{Cls} groups families as in Sections~\ref{sec:corr}--\ref{sec:temporal}: C~= correlation/iterative, P~= classical\,/\,pyramid, E~= efficient\,/\,reduced-supervision, T~= transformer\,/\,global matching, M~= temporal\,/\,memory\,/\,generative. $\dagger$~marks models consuming more than two input frames. The last row is the zero-flow reference (predicting $\mathbf{0}$ everywhere).}
  \label{tab:summary-best-drone-270x480}
  \centering
  \small
  \resizebox{\linewidth}{!}{%
  \begin{tabular}{cllrrrrrr}
    \toprule
    Cls & Model & Checkpoint & EPE [95\% CI] & EPE$_\text{n}$ & Px1 & Px3 & Px5 & Fl \\
    \midrule
    E & ARFlow & KITTI12/pwclite\_ar\_mv & 0.873 {\scriptsize[0.77,0.99]} & 0.215 & 19.95 & 6.779 & 3.543 & 6.779 \\
    M & MemFlow$^\dagger$ & MemFlowNet\_spring & 0.948 {\scriptsize[0.85,1.05]} & 0.235 & 23.05 & 7.776 & 3.343 & 7.776 \\
    T & UniMatch & gmflow-scale2-regrefine6-mixdata-train320x576-4e7b215d &0.952 {\scriptsize[0.86,1.05]} & 0.237 & 24.55 & 7.568 & 3.310 & 7.568 \\
    P & PyFlow (C2F) & fast & 0.977 {\scriptsize[0.85,1.11]} & 0.240 & 18.53 & 7.751 & 4.785 & 7.751 \\
    M & VideoFlow$^\dagger$ & MOF\_things\_288960noise & 0.997 {\scriptsize[0.91,1.09]} & 0.251 & 26.17 & 8.343 & 3.065 & 8.343 \\
    C & WAFT & tar-c-t-spring-540p & 1.000 {\scriptsize[0.89,1.11]} & 0.245 & 24.82 & 8.815 & 3.802 & 8.815 \\
    P & VCN & vcn\_weights/things/finetune\_211999 & 1.067 {\scriptsize[0.96,1.18]} & 0.263 & 27.17 & 8.788 & 4.077 & 8.788 \\
    M & FlowDiffusion (DDVM) & pipeline-900000 & 1.099 {\scriptsize[1.00,1.21]} & 0.272 & 29.51 & 9.322 & 3.843 & 9.322 \\
    T & FlowFormer++ & things & 1.108 {\scriptsize[1.01,1.21]} & 0.277 & 29.02 & 9.245 & 3.787 & 9.245 \\
    C & SEA-RAFT & Tartan-C-T-TSKH432x960-M (sintel) & 1.117 {\scriptsize[1.02,1.22]} & 0.278 & 30.13 & 9.349 & 3.996 & 9.349 \\
    C & RAFT & raft-things & 1.152 {\scriptsize[1.05,1.26]} & 0.288 & 31.43 & 9.854 & 3.854 & 9.854 \\
    P & LiteFlowNet3 & liteflownet3-sintel & 1.192 {\scriptsize[1.08,1.31]} & 0.297 & 30.89 & 10.60 & 4.448 & 10.60 \\
    P & FlowNet2 & FlowNet2\_checkpoint & 1.209 {\scriptsize[1.10,1.33]} & 0.300 & 31.52 & 10.30 & 4.625 & 10.30 \\
    P & PWC-Net & pwc\_net\_chairs & 1.227 {\scriptsize[1.11,1.35]} & 0.306 & 32.60 & 10.30 & 4.317 & 10.30 \\
    T & GMFlow & gmflow\_with\_refine\_things-36579974 & 1.282 {\scriptsize[1.17,1.39]} & 0.313 & 32.96 & 10.19 & 4.194 & 10.19 \\
    E & FastFlowNet & fastflownet\_ft\_kitti & 1.459 {\scriptsize[1.32,1.60]} & 0.361 & 36.35 & 13.73 & 6.257 & 13.73 \\
    \midrule
    -- & Zero flow (reference) & -- & 1.744 {\scriptsize[1.61,1.89]} & 0.417 & 47.42 & 16.08 & 7.084 & 16.08 \\
    \bottomrule
  \end{tabular}%
  }
\end{table}

%% file: tables/tab_summary_best_drone_432x960.tex
\begin{table}[h]
  \caption{Best drone-region EPE per model family at 960$\times$432, on the stratified subset (999 flow pairs). One row per family: the checkpoint minimising drone-region EPE. Brackets give the 95\% percentile-bootstrap confidence interval of the drone-region EPE (10{,}000 resamples over pairs). \emph{Cls} groups families as in Sections~\ref{sec:corr}--\ref{sec:temporal}: C~= correlation/iterative, P~= classical\,/\,pyramid, E~= efficient\,/\,reduced-supervision, T~= transformer\,/\,global matching, M~= temporal\,/\,memory\,/\,generative. $\dagger$~marks models consuming more than two input frames. The last row is the zero-flow reference (predicting $\mathbf{0}$ everywhere).}
  \label{tab:summary-best-drone-432x960}
  \centering
  \small
  \resizebox{\linewidth}{!}{%
  \begin{tabular}{cllrrrrrr}
    \toprule
    Cls & Model & Checkpoint & EPE [95\% CI] & EPE$_\text{n}$ & Px1 & Px3 & Px5 & Fl \\
    \midrule
    T & UniMatch & gmflow-scale2-regrefine6-things-776ed612 & 1.293 {\scriptsize[1.18,1.41]} & 0.175 & 26.70 & 11.60 & 5.926 & 11.60 \\
    M & VideoFlow$^\dagger$ & BOF\_sintel & 1.425 {\scriptsize[1.31,1.55]} & 0.192 & 30.82 & 13.16 & 6.967 & 13.16 \\
    M & MemFlow$^\dagger$ & MemFlowNet\_spring & 1.484 {\scriptsize[1.32,1.66]} & 0.192 & 27.03 & 12.60 & 7.300 & 12.60 \\
    T & FlowFormer++ & things\_288960 & 1.513 {\scriptsize[1.37,1.66]} & 0.205 & 33.06 & 13.41 & 7.099 & 13.41 \\
    C & SEA-RAFT & Tartan-C-T-TSKH432x960-M (sintel) & 1.557 {\scriptsize[1.40,1.72]} & 0.204 & 30.26 & 14.48 & 7.701 & 14.48 \\
    E & ARFlow & KITTI12/pwclite\_ar\_mv & 1.581 {\scriptsize[1.36,1.82]} & 0.199 & 24.74 & 12.14 & 7.919 & 12.14 \\
    T & GMFlow & gmflow\_with\_refine\_sintel-3ed1cf48 & 1.665 {\scriptsize[1.52,1.81]} & 0.222 & 34.63 & 15.63 & 8.260 & 15.63 \\
    C & RAFT & raft-sintel & 1.713 {\scriptsize[1.55,1.89]} & 0.228 & 35.03 & 15.10 & 8.106 & 15.10 \\
    C & WAFT & tar-c-t-spring-540p & 1.793 {\scriptsize[1.58,2.02]} & 0.228 & 30.85 & 15.88 & 9.637 & 15.88 \\
    P & PyFlow (C2F) & fast & 1.895 {\scriptsize[1.62,2.20]} & 0.243 & 23.33 & 13.80 & 9.455 & 13.80 \\
    P & FlowNet2 & FlowNet2-CS\_checkpoint & 2.011 {\scriptsize[1.81,2.22]} & 0.258 & 40.65 & 17.62 & 10.28 & 17.62 \\
    P & VCN & vcn\_weights/sintel-ft-trainval/finetune\_67999 & 2.048 {\scriptsize[1.84,2.26]} & 0.262 & 39.78 & 18.74 & 10.93 & 18.74 \\
    P & PWC-Net & pwc\_net\_chairs & 2.125 {\scriptsize[1.91,2.35]} & 0.278 & 44.61 & 18.30 & 10.40 & 18.30 \\
    P & LiteFlowNet3 & liteflownet3-sintel & 2.147 {\scriptsize[1.92,2.39]} & 0.276 & 37.79 & 19.55 & 11.60 & 19.55 \\
    M & FlowDiffusion (DDVM) & pipeline-305000 & 2.220 {\scriptsize[2.00,2.45]} & 0.284 & 41.94 & 21.23 & 11.96 & 21.23 \\
    E & FastFlowNet & fastflownet\_things3d & 2.336 {\scriptsize[2.08,2.60]} & 0.296 & 40.44 & 21.24 & 12.36 & 21.24 \\
    \midrule
    -- & Zero flow (reference) & -- & 3.498 {\scriptsize[3.23,3.78]} & 0.432 & 72.33 & 36.55 & 21.50 & 36.55 \\
    \bottomrule
  \end{tabular}%
  }
\end{table}

%% file: tables/tab_summary_best_drone_1080x1920.tex
\begin{table}[h]
  \caption{Best drone-region EPE per model family at 1920$\times$1080, on the stratified subset (999 flow pairs). One row per family: the checkpoint minimising drone-region EPE. Brackets give the 95\% percentile-bootstrap confidence interval of the drone-region EPE (10{,}000 resamples over pairs). \emph{Cls} groups families as in Sections~\ref{sec:corr}--\ref{sec:temporal}: C~= correlation/iterative, P~= classical\,/\,pyramid, E~= efficient\,/\,reduced-supervision, T~= transformer\,/\,global matching, M~= temporal\,/\,memory\,/\,generative. $\dagger$~marks models consuming more than two input frames; $\ddagger$~marks runs that used tiled inference ($960\times432$ tiles; $960\times288$ for FlowFormer++ \texttt{things\_288960}). The last row is the zero-flow reference (predicting $\mathbf{0}$ everywhere).}
  \label{tab:summary-best-drone-1080x1920}
  \centering
  \small
  \resizebox{\linewidth}{!}{%
\begin{tabular}{cllrrrrrr}
    \toprule
    Cls & Model & Checkpoint & EPE [95\% CI] & EPE$_\text{n}$ & Px1 & Px3 & Px5 & Fl \\
    \midrule
    T & UniMatch$^\ddagger$ & gmflow-scale2-regrefine6-sintelft-6e39e2b9 & 1.557 {\scriptsize[1.42,1.70]} & 0.097 & 25.26 & 12.12 & 7.931 & 12.12 \\
    M & VideoFlow$^\dagger$$^\ddagger$ & BOF\_sintel & 1.582 {\scriptsize[1.45,1.72]} & 0.099 & 26.66 & 12.06 & 8.111 & 12.05 \\
    T & FlowFormer++$^\ddagger$ & things\_288960 & 1.840 {\scriptsize[1.67,2.02]} & 0.115 & 29.26 & 13.92 & 9.244 & 13.92 \\
    C & SEA-RAFT & Tartan-C-T-TSKH432x960-M (sintel) & 1.883 {\scriptsize[1.70,2.08]} & 0.115 & 27.80 & 14.37 & 10.14 & 14.37 \\
    M & MemFlow$^\dagger$$^\ddagger$ & MemFlowNet\_T\_sintel & 1.965 {\scriptsize[1.77,2.18]} & 0.120 & 29.76 & 15.17 & 10.64 & 15.17 \\
    T & GMFlow$^\ddagger$ & gmflow\_with\_refine\_sintel-3ed1cf48 & 1.996 {\scriptsize[1.82,2.19]} & 0.124 & 32.64 & 15.65 & 9.978 & 15.65 \\
    C & RAFT$^\ddagger$ & raft-sintel & 2.047 {\scriptsize[1.86,2.24]} & 0.127 & 34.04 & 14.97 & 9.926 & 14.95 \\
\bottomrule
  \end{tabular}%
  }
\end{table}

%% file: tables/tab_agg_full.tex
\begingroup
\setlength{\LTleft}{0pt}
\setlength{\LTright}{0pt}
\setlength{\tabcolsep}{3.3pt}
\scriptsize
\renewcommand{\arraystretch}{1.08}
\begin{longtable}{>{\raggedright\arraybackslash}p{2.25cm}>{\raggedright\arraybackslash}p{4.2cm}>{\raggedright\arraybackslash}p{1.7cm}rrrrrr}
\caption{Full-frame evaluation results on FlyingDrones, all model families, on the stratified subset (999 flow pairs; each reference frame $t$ is paired with $t-1$). Rows grouped by model family, resolution, then checkpoint. \textit{Italic}: best result within the model family at that resolution (lowest full-frame EPE). EPE is end-point error; EPE$_\text{n}$ is the resolution-normalized EPE; Px$k$ is the percentage of pixels with error $>k$; Fl is the outlier rate (within 0.02 percentage points of Px3 here). $\dagger$~marks the two families consuming more than two input frames: VideoFlow (\texttt{BOF\_*} three frames, \texttt{MOF\_*} five) and MemFlow (recurrent memory over an extended window). $\ddagger$~marks $1920\times1080$ runs that used tiled inference ($960\times432$ tiles; $960\times288$ for FlowFormer++ \texttt{things\_288960}). The three SEA-RAFT rows are the separately released \texttt{kitti}, \texttt{sintel} and \texttt{spring} checkpoints. The last block is the zero-flow reference (predicting $\mathbf{0}$ everywhere).}\label{tab:agg-full}\\
\toprule
\textbf{Model} & \textbf{Checkpoint} & \textbf{Res.} & \textbf{EPE} & \textbf{EPE$_\text{n}$} & \textbf{Px1} & \textbf{Px3} & \textbf{Px5} & \textbf{Fl} \\
\midrule
\endfirsthead

\multicolumn{9}{l}{\textit{Table~\ref{tab:agg-full} continued from previous page.}} \\
\toprule
\textbf{Model} & \textbf{Checkpoint} & \textbf{Res.} & \textbf{EPE} & \textbf{EPE$_\text{n}$} & \textbf{Px1} & \textbf{Px3} & \textbf{Px5} & \textbf{Fl} \\
\midrule
\endhead

\midrule
\multicolumn{9}{r}{\textit{Continued on next page.}} \\
\endfoot

\bottomrule
\endlastfoot

  ARFlow & CityScapes/pwclite\_ar & 480$\times$270 & 0.331 & 0.091 & 7.451 & 0.645 & 0.025 & 0.645 \\
  ARFlow & KITTI12/pwclite\_ar & 480$\times$270 & 0.168 & 0.048 & 2.342 & 0.177 & 0.064 & 0.177 \\
  \textit{ARFlow} & \textit{KITTI12/pwclite\_ar\_mv} & \textit{480$\times$270} & \textit{0.150} & \textit{0.042} & \textit{2.371} & \textit{0.128} & \textit{0.025} & \textit{0.128} \\
  ARFlow & KITTI15/pwclite\_ar & 480$\times$270 & 0.189 & 0.051 & 2.962 & 0.513 & 0.256 & 0.513 \\
  ARFlow & KITTI15/pwclite\_ar\_mv & 480$\times$270 & 0.157 & 0.043 & 2.373 & 0.273 & 0.112 & 0.273 \\
  ARFlow & KITTI15/pwclite\_raw & 480$\times$270 & 0.225 & 0.059 & 2.524 & 0.371 & 0.271 & 0.371 \\
  ARFlow & Sintel/pwclite\_ar & 480$\times$270 & 0.234 & 0.063 & 5.846 & 0.342 & 0.017 & 0.342 \\
  ARFlow & Sintel/pwclite\_ar\_mv & 480$\times$270 & 0.218 & 0.061 & 5.283 & 0.567 & 0.017 & 0.567 \\
\end{longtable}
\endgroup

%% file: tables/tab_agg_drone.tex
\begingroup
\setlength{\LTleft}{0pt}
\setlength{\LTright}{0pt}
\setlength{\tabcolsep}{3.3pt}
\scriptsize
\renewcommand{\arraystretch}{1.08}
\begin{longtable}{>{\raggedright\arraybackslash}p{2.25cm}>{\raggedright\arraybackslash}p{4.2cm}>{\raggedright\arraybackslash}p{1.7cm}rrrrrr}
\caption{Drone-region (drone-pixel-only) evaluation results on FlyingDrones, all model families, on the stratified subset (999 flow pairs; each reference frame $t$ is paired with $t-1$). Rows grouped by model family, resolution, then checkpoint. \textit{Italic}: best result within the model family at that resolution (lowest drone-region EPE). EPE is end-point error; EPE$_\text{n}$ is the resolution-normalized EPE; Px$k$ is the percentage of pixels with error $>k$; Fl is the outlier rate (within 0.02 percentage points of Px3 here). $\dagger$~marks the two families consuming more than two input frames: VideoFlow (\texttt{BOF\_*} three frames, \texttt{MOF\_*} five) and MemFlow (recurrent memory over an extended window). $\ddagger$~marks $1920\times1080$ runs that used tiled inference ($960\times432$ tiles; $960\times288$ for FlowFormer++ \texttt{things\_288960}). The three SEA-RAFT rows are the separately released \texttt{kitti}, \texttt{sintel} and \texttt{spring} checkpoints. The last block is the zero-flow reference (predicting $\mathbf{0}$ everywhere).}\label{tab:agg-drone}\\
\toprule
\textbf{Model} & \textbf{Checkpoint} & \textbf{Res.} & \textbf{EPE} & \textbf{EPE$_\text{n}$} & \textbf{Px1} & \textbf{Px3} & \textbf{Px5} & \textbf{Fl} \\
\midrule
\endfirsthead

\multicolumn{9}{l}{\textit{Table~\ref{tab:agg-drone} continued from previous page.}} \\
\toprule
\textbf{Model} & \textbf{Checkpoint} & \textbf{Res.} & \textbf{EPE} & \textbf{EPE$_\text{n}$} & \textbf{Px1} & \textbf{Px3} & \textbf{Px5} & \textbf{Fl} \\
\midrule
\endhead

\midrule
\multicolumn{9}{r}{\textit{Continued on next page.}} \\
\endfoot

\bottomrule
\endlastfoot

  ARFlow & CityScapes/pwclite\_ar & 480$\times$270 & 0.940 & 0.231 & 21.28 & 7.538 & 3.650 & 7.538 \\
  ARFlow & KITTI12/pwclite\_ar & 480$\times$270 & 0.896 & 0.222 & 20.74 & 6.834 & 3.433 & 6.834 \\
  \textit{ARFlow} & \textit{KITTI12/pwclite\_ar\_mv} & \textit{480$\times$270} & \textit{0.873} & \textit{0.215} & \textit{19.95} & \textit{6.779} & \textit{3.543} & \textit{6.779} \\
  ARFlow & KITTI15/pwclite\_ar & 480$\times$270 & 0.897 & 0.222 & 20.12 & 7.009 & 3.501 & 7.009 \\
  ARFlow & KITTI15/pwclite\_ar\_mv & 480$\times$270 & 0.886 & 0.217 & 19.92 & 7.011 & 3.632 & 7.011 \\
  ARFlow & KITTI15/pwclite\_raw & 480$\times$270 & 0.896 & 0.222 & 20.61 & 6.864 & 3.448 & 6.864 \\
  ARFlow & Sintel/pwclite\_ar & 480$\times$270 & 1.055 & 0.256 & 23.49 & 9.106 & 4.711 & 9.106 \\
  ARFlow & Sintel/pwclite\_ar\_mv & 480$\times$270 & 1.001 & 0.242 & 21.89 & 8.636 & 4.282 & 8.636 \\
\end{longtable}
\endgroup

%% file: tables/tab_protocol.tex
\begin{table}[t]
\centering
\caption{Drone-region EPE for the leaderboard checkpoint of each family, contrasting a contiguous window (reference frames 1--500, each paired with its predecessor) against the stratified subset (999 pairs) used throughout the paper (checkpoint held fixed). \emph{Window} is the contiguous window; \emph{Strat.} is the stratified subset. $\Delta$ is Strat.\ $-$ Window. Median (mean) $|\Delta|$: 0.69 (0.69)\,px at $480\times270$, 1.18 (1.15)\,px at $960\times432$, 1.29 (1.64)\,px at $1920\times1080$. The last row of each block is the zero-flow reference.}
\label{tab:protocol}
\small
\setlength{\tabcolsep}{5pt}
\resizebox{\linewidth}{!}{%
\begin{tabular}{llrrr@{\hskip 1.2em}llrrr}
\toprule
\multicolumn{5}{c}{$480\times270$} & \multicolumn{5}{c}{$960\times432$} \\
\cmidrule(lr){1-5}\cmidrule(lr){6-10}
Model & Ckpt & Window & Strat. & $\Delta$ & Model & Ckpt & Window & Strat. & $\Delta$ \\
\midrule
PyFlow & \texttt{fast} & 1.395 & 0.977 & -0.42 & VideoFlow & \texttt{BOF\_sintel} & 1.797 & 1.425 & -0.37 \\
UniMatch & \texttt{scale2-regrefine6-mixdata-train320x576} & 1.472 & 0.952 & -0.52 & UniMatch & \texttt{scale2-regrefine6-things} & 1.954 & 1.293 & -0.66 \\
ARFlow & \texttt{KITTI12/pwclite\_ar\_mv} & 1.516 & 0.873 & -0.64 & GMFlow & \texttt{gmflow\_with\_refine\_sintel} & 2.154 & 1.665 & -0.49 \\
MemFlow & \texttt{MemFlowNet\_spring} & 1.634 & 0.948 & -0.69 & FlowFormer++ & \texttt{things\_288960} & 2.438 & 1.513 & -0.93 \\
VideoFlow & \texttt{MOF\_things\_288960noise} & 1.641 & 0.997 & -0.64 & SEA-RAFT & \texttt{TSKH432x960-M (sintel)} & 2.472 & 1.557 & -0.92 \\
SEA-RAFT & \texttt{TSKH432x960-M (sintel)} & 1.774 & 1.117 & -0.66 & MemFlow & \texttt{MemFlowNet\_spring} & 2.538 & 1.484 & -1.05 \\
VCN & \texttt{things/211999} & 1.793 & 1.067 & -0.73 & RAFT & \texttt{raft-sintel} & 2.740 & 1.713 & -1.03 \\
FlowFormer++ & \texttt{things} & 1.797 & 1.108 & -0.69 & PyFlow & \texttt{fast} & 2.742 & 1.895 & -0.85 \\
GMFlow & \texttt{gmflow\_with\_refine\_things} & 1.823 & 1.282 & -0.54 & ARFlow & \texttt{KITTI12/pwclite\_ar\_mv} & 2.882 & 1.581 & -1.30 \\
WAFT & \texttt{tar-c-t-spring-540p} & 1.832 & 1.000 & -0.83 & WAFT & \texttt{tar-c-t-spring-540p} & 3.332 & 1.793 & -1.54 \\
FlowDiffusion & \texttt{pipeline-900000} & 1.893 & 1.099 & -0.79 & FlowNet2 & \texttt{FlowNet2-CS} & 3.361 & 2.011 & -1.35 \\
FlowNet2 & \texttt{FlowNet2} & 1.895 & 1.209 & -0.69 & VCN & \texttt{sintel-ft-trainval/67999} & 3.420 & 2.048 & -1.37 \\
PWC-Net & \texttt{pwc\_net\_chairs} & 1.957 & 1.227 & -0.73 & PWC-Net & \texttt{pwc\_net\_chairs} & 3.611 & 2.125 & -1.49 \\
RAFT & \texttt{raft-things} & 2.017 & 1.152 & -0.87 & FastFlowNet & \texttt{things3d} & 3.898 & 2.336 & -1.56 \\
LiteFlowNet3 & \texttt{liteflownet3-sintel} & 2.052 & 1.192 & -0.86 & LiteFlowNet3 & \texttt{liteflownet3-sintel} & 3.907 & 2.147 & -1.76 \\
FastFlowNet & \texttt{ft\_kitti} & 2.193 & 1.459 & -0.73 & FlowDiffusion & \texttt{pipeline-305000} & 3.925 & 2.220 & -1.70 \\
\midrule
Zero flow & -- & 2.229 & 1.744 & -0.48 & Zero flow & -- & 4.356 & 3.498 & -0.86 \\
\bottomrule
\end{tabular}%
}

\vspace{0.8em}
\begin{tabular}{llrrr}
\toprule
\multicolumn{5}{c}{$1920\times1080$} \\
\midrule
Model & Ckpt & Window & Strat. & $\Delta$ \\
\midrule
UniMatch & \texttt{scale2-regrefine6-sintelft} & 1.889 & 1.557 & -0.33 \\
VideoFlow & \texttt{BOF\_sintel} & 1.952 & 1.582 & -0.37 \\
MemFlow & \texttt{MemFlowNet\_T\_sintel} & 2.206 & 1.965 & -0.24 \\
GMFlow & \texttt{gmflow\_with\_refine\_sintel} & 2.219 & 1.996 & -0.22 \\
SEA-RAFT & \texttt{TSKH432x960-M (sintel)} & 2.541 & 1.883 & -0.66 \\
FlowFormer++ & \texttt{things\_288960} & 2.621 & 1.840 & -0.78 \\
RAFT & \texttt{raft-sintel} & 3.056 & 2.047 & -1.01 \\
WAFT & \texttt{tar-c-t} & 3.274 & 2.086 & -1.19 \\
LiteFlowNet3 & \texttt{liteflownet3-sintel} & 5.286 & 3.275 & -2.01 \\
FlowNet2 & \texttt{FlowNet2} & 5.346 & 3.136 & -2.21 \\
PyFlow & \texttt{fast} & 5.438 & 4.051 & -1.39 \\
ARFlow & \texttt{KITTI15/pwclite\_ar} & 6.046 & 3.175 & -2.87 \\
VCN & \texttt{chairs/141999} & 6.329 & 3.829 & -2.50 \\
PWC-Net & \texttt{pwc\_net} & 7.553 & 4.404 & -3.15 \\
FastFlowNet & \texttt{ft\_kitti} & 7.602 & 4.425 & -3.18 \\
FlowDiffusion & \texttt{pipeline-305000} & 9.181 & 5.125 & -4.06 \\
\midrule
Zero flow & -- & 9.655 & 7.549 & -2.11 \\
\bottomrule
\end{tabular}
\end{table}

%% file: tables/tab_real_all.tex
\begin{table}[t]
\centering
\caption{Real-footage motion-alignment metrics for all 41 published checkpoints, mean $\pm$ sample standard deviation (ddof\,=\,1) over all 21 labelled cameras of datasets 1--4 (100 displacement-stratified frame pairs per camera), ordered by drone-disk NSS. Inference at $1280\times720$ (VideoFlow MOF at $960\times540$), every field rescaled to $1920\times1080$. GMFlow, UniMatch and FlowFormer++ exceed their $960\times432$ tile size at this resolution (FlowFormer++ \texttt{things\_288960}: $960\times288$) and are run with overlapping tiles blended with uniform weights; all other models run in a single pass. UniMatch \texttt{gmflow-scale1-things} and GMFlow \texttt{gmflow\_things} share identical weights and hence identical scores. VCN paths are relative to \texttt{vcn\_weights/} (\texttt{small/} = \texttt{vcn\_small\_weights/}). Higher is better except pt-EPE.}
\label{tab:real-all}
\small
\setlength{\tabcolsep}{4pt}
\resizebox{\linewidth}{!}{%
\begin{tabular}{llrrrrrr}
\toprule
Family & Checkpoint & $n$ & $\mathrm{NSS}_D$ & Hit@50 & Contr. & cos & pt-EPE \\
\midrule
MemFlow       & \texttt{MemFlowNet\_spring} & 21 &  27.3$\pm$10.2 & 0.62 &  303.8 &  0.60 &  1.33 \\
PyFlow        & \texttt{fast} & 21 &  26.1$\pm$ 9.9 & 0.71 &   74.6 &  0.79 &  1.01 \\
SEA-RAFT      & \texttt{Tartan-C-T-TSKH432x960-M (sintel)} & 21 &  16.5$\pm$ 8.9 & 0.45 &   35.4 &  0.46 &  1.52 \\
MemFlow       & \texttt{MemFlowNet\_sintel} & 21 &  16.3$\pm$ 7.7 & 0.46 &   42.8 &  0.34 &  1.80 \\
RAFT          & \texttt{raft-sintel} & 21 &  15.5$\pm$ 8.4 & 0.40 &   23.7 &  0.41 &  1.75 \\
FlowFormer++  & \texttt{things\_288960}$^\ddagger$ & 21 &  15.5$\pm$ 8.7 & 0.56 &   12.8 &  0.54 &  1.58 \\
MemFlow       & \texttt{MemFlowNet\_things} & 21 &  15.4$\pm$ 7.6 & 0.49 &   18.7 &  0.34 &  1.74 \\
MemFlow       & \texttt{MemFlowNet\_kitti} & 21 &  15.1$\pm$ 8.4 & 0.46 &   51.5 &  0.31 &  1.66 \\
RAFT          & \texttt{raft-things} & 21 &  14.5$\pm$ 8.7 & 0.52 &   14.4 &  0.48 &  1.62 \\
ARFlow        & \texttt{Sintel/pwclite\_ar} & 21 &  13.5$\pm$ 5.9 & 0.54 &   14.4 &  0.80 &  1.19 \\
UniMatch      & \texttt{gmflow-scale2-regrefine6-kitti15-25b554d7}$^\ddagger$ & 21 &  13.3$\pm$ 9.0 & 0.46 &   31.0 &  0.47 &  2.06 \\
FlowFormer++  & \texttt{sintel}$^\ddagger$ & 21 &  12.0$\pm$ 6.6 & 0.41 &    9.9 &  0.40 &  1.90 \\
GMFlow        & \texttt{gmflow\_with\_refine\_kitti-8d3b9786}$^\ddagger$ & 21 &   8.9$\pm$ 5.2 & 0.34 &   33.3 &  0.47 &  2.05 \\
RAFT          & \texttt{raft-chairs} & 21 &   8.8$\pm$ 8.1 & 0.24 &   14.6 &  0.22 &  2.18 \\
RAFT          & \texttt{raft-kitti} & 21 &   8.6$\pm$ 7.2 & 0.28 &   40.5 &  0.15 &  1.83 \\
VCN           & \texttt{sintel-ft-trainval/finetune\_67999} & 21 &   8.5$\pm$ 5.2 & 0.32 &    7.7 &  0.76 &  1.46 \\
VCN           & \texttt{chairs/finetune\_141999} & 21 &   8.3$\pm$ 4.6 & 0.37 &    6.3 &  0.71 &  1.78 \\
VCN           & \texttt{things/finetune\_239999} & 21 &   7.7$\pm$ 4.2 & 0.39 &    5.8 &  0.77 &  1.41 \\
LiteFlowNet3  & \texttt{liteflownet3s-sintel} & 21 &   7.1$\pm$ 5.2 & 0.30 &    6.4 &  0.69 &  1.69 \\
GMFlow        & \texttt{gmflow\_kitti-285701a8}$^\ddagger$ & 21 &   6.9$\pm$ 4.9 & 0.30 &   27.8 &  0.38 &  2.15 \\
UniMatch      & \texttt{gmflow-scale1-things-e9887eda}$^\ddagger$ & 21 &   6.8$\pm$ 5.4 & 0.31 &    7.0 &  0.33 &  2.23 \\
GMFlow        & \texttt{gmflow\_things-e9887eda}$^\ddagger$ & 21 &   6.8$\pm$ 5.4 & 0.31 &    7.0 &  0.33 &  2.23 \\
FastFlowNet   & \texttt{fastflownet\_ft\_kitti} & 21 &   6.6$\pm$ 3.9 & 0.35 &    6.7 &  0.66 &  1.46 \\
GMFlow        & \texttt{gmflow\_sintel-0c07dcb3}$^\ddagger$ & 21 &   6.6$\pm$ 5.4 & 0.30 &    9.1 &  0.26 &  2.38 \\
GMFlow        & \texttt{gmflow\_with\_refine\_chairs-020cc9be}$^\ddagger$ & 21 &   6.2$\pm$ 4.9 & 0.26 &    5.3 &  0.45 &  2.16 \\
WAFT          & \texttt{tar-c-t-kitti} & 21 &   6.0$\pm$ 4.8 & 0.27 &    7.2 &  0.34 &  1.84 \\
VCN           & \texttt{kitti-ft-trainval/finetune\_149999} & 21 &   5.8$\pm$ 3.0 & 0.37 &    5.0 &  0.72 &  1.22 \\
VCN           & \texttt{small/chairs/finetune\_139999} & 21 &   5.7$\pm$ 3.7 & 0.23 &    4.4 &  0.62 &  1.94 \\
FlowFormer++  & \texttt{kitti}$^\ddagger$ & 21 &   5.5$\pm$ 4.0 & 0.24 &   13.1 &  0.18 &  2.11 \\
FastFlowNet   & \texttt{fastflownet\_things3d} & 21 &   5.5$\pm$ 3.1 & 0.34 &    4.6 &  0.69 &  1.47 \\
VideoFlow     & \texttt{MOF\_sintel} (MOF) & 21 &   4.9$\pm$ 4.4 & 0.16 &    5.1 &  0.12 &  2.25 \\
PWC-Net       & \texttt{pwc\_net} & 21 &   4.9$\pm$ 4.0 & 0.26 &    4.5 &  0.53 &  2.16 \\
FastFlowNet   & \texttt{fastflownet\_ft\_sintel} & 21 &   4.8$\pm$ 3.1 & 0.26 &    4.1 &  0.59 &  1.72 \\
PWC-Net       & \texttt{pwc\_net\_chairs} & 21 &   4.6$\pm$ 2.7 & 0.26 &    3.9 &  0.62 &  1.81 \\
GMFlow        & \texttt{gmflow\_chairs-1d776046}$^\ddagger$ & 21 &   4.4$\pm$ 4.0 & 0.23 &    4.3 &  0.18 &  2.41 \\
WAFT          & \texttt{tar-c-t-sintel} & 21 &   3.9$\pm$ 4.6 & 0.15 &    6.8 &  0.31 &  2.27 \\
LiteFlowNet3  & \texttt{liteflownet3s-kitti} & 21 &   3.9$\pm$ 3.3 & 0.19 &    3.8 &  0.49 &  1.76 \\
FastFlowNet   & \texttt{fastflownet\_chairs} & 21 &   3.7$\pm$ 2.5 & 0.25 &    3.4 &  0.64 &  1.74 \\
FlowDiffusion & \texttt{pipeline-305000} & 21 &   2.9$\pm$ 2.8 & 0.14 &    2.5 &  0.22 &  2.33 \\
FlowNet2      & \texttt{FlowNet2-C\_checkpoint} & 21 &   1.5$\pm$ 1.5 & 0.12 &    1.8 &  0.37 &  2.17 \\
FlowNet2      & \texttt{FlowNet2-S\_checkpoint} & 21 &   1.0$\pm$ 1.3 & 0.08 &    1.5 &  0.31 &  2.36 \\
\bottomrule
\end{tabular}%
}
\par\smallskip{\footnotesize $^\ddagger$\,tiled inference ($960\times432$ tiles, $960\times288$ for \texttt{things\_288960}).}
\end{table}

%% file: tables/tab_real_strata.tex
\begin{table}[t]
\centering
\caption{Localisation (Hit@50) versus drone image-motion stratum for representative checkpoints, mean over the labelled cameras that contain pairs in the stratum (21 cameras for sub-px, low and mid; 18 for high). Localisation rises with motion for the transferable models, including RAFT; the one failing checkpoint shown here (VideoFlow \texttt{sintel}) stays low at every level.}
\label{tab:real-strata}
\small
\setlength{\tabcolsep}{6pt}
\begin{tabular}{lrrrr}
\toprule
Checkpoint & sub-px & low & mid & high \\
 & ($<1$\,px) & ($1$--$3$) & ($3$--$8$) & ($>8$) \\
\midrule
MemFlow / \texttt{spring}              & 0.24 & 0.63 & 0.86 & 0.90 \\
PyFlow / \texttt{fast}                 & 0.54 & 0.74 & 0.81 & 0.88 \\
FlowFormer++ / \texttt{things\_288960} & 0.22 & 0.48 & 0.84 & 0.91 \\
GMFlow / \texttt{with\_refine\_kitti}  & 0.15 & 0.22 & 0.47 & 0.85 \\
VideoFlow MOF / \texttt{sintel}        & 0.07 & 0.11 & 0.25 & 0.39 \\
RAFT / \texttt{sintel}                 & 0.08 & 0.25 & 0.71 & 0.89 \\
\bottomrule
\end{tabular}
\end{table}

%% file: tables/tab_ft.tex
\begin{table}[t]
\centering
\caption{Fine-tuning recipe and optimisation behaviour per arm (800-pair FlyingDrones training split, 1500 AdamW steps, one-cycle schedule, mixed precision; frames resized to the evaluation scale, then randomly cropped). \emph{Batch} is per-step batch $\times$ gradient-accumulation steps; $\gamma$ is the sequence-loss weight decay. \emph{Val drone-EPE} is the canonical drone-region EPE on the 100-pair validation split at step~0 (public initialisation), at the selected checkpoint, and at step 1500; validation runs every 150 steps and the checkpoint minimising it (\emph{Best}: its step) is reported.}
\label{tab:ft-curves}
\small
\setlength{\tabcolsep}{4pt}
\resizebox{\linewidth}{!}{%
\begin{tabular}{llcccccrrrr}
\toprule
 & & & & & & & & \multicolumn{3}{c}{Val drone-EPE} \\
\cmidrule(lr){9-11}
Model & Init & Eval res. & Crop & Batch & LR & $\gamma$ & Best & step 0 & best & 1500 \\
\midrule
FlowFormer++ & \texttt{things\_288960} & $960\times432$ & $768\times384$ & $1\times6$ & $1.25\times10^{-4}$ & 0.85 & 600 & 1.042 & 0.986 & 0.989 \\
GMFlow & \texttt{gmflow\_chairs} & $1280\times720$ & $896\times512$ & $1\times1$ & $5.0\times10^{-5}$ & 0.90 & 300 & 2.239 & 1.856 & 1.967 \\
RAFT & \texttt{raft-things} & $960\times432$ & $768\times384$ & $4\times1$ & $2.5\times10^{-5}$ & 0.85 & 150 & 1.105 & 1.107 & 1.195 \\
VideoFlow MOF & \texttt{MOF\_sintel} & $960\times432$ & $768\times384$ & $1\times4$ & $6.25\times10^{-5}$ & 0.85 & 1500 & 1.116 & 0.840 & 0.840 \\
\bottomrule
\end{tabular}%
}
\end{table}

\begin{table}[t]
\centering
\caption{Fine-tuning results, public initialisation (base) $\rightarrow$ fine-tuned (best validation checkpoint). \emph{Synthetic}: held-out FlyingDrones test split (100 pairs never seen in training or validation) at the arm's evaluation resolution, canonical evaluator; drone-region and full-frame EPE. \emph{Real}: the same two checkpoints on the real-footage motion-alignment protocol of Section~\ref{sec:real} (21 cameras, mean; $1280\times720$, VideoFlow MOF $960\times540$).}
\label{tab:ft-test}
\small
\setlength{\tabcolsep}{4pt}
\resizebox{\linewidth}{!}{%
\begin{tabular}{lcrrrrr}
\toprule
 & & \multicolumn{3}{c}{Synthetic test (FlyingDrones)} & \multicolumn{2}{c}{Real footage} \\
\cmidrule(lr){3-5}\cmidrule(lr){6-7}
Model & Res. & Drone EPE & $\Delta$ & Full EPE & $\mathrm{NSS}_D$ & Hit@50 \\
\midrule
FlowFormer++ & $960\times432$ & $1.614\rightarrow1.305$ & $-19.2\%$ & $0.619\rightarrow0.072$ & $15.5\rightarrow22.6$ & $0.56\rightarrow0.58$ \\
GMFlow & $1280\times720$ & $3.649\rightarrow2.723$ & $-25.4\%$ & $1.394\rightarrow0.794$ & $4.4\rightarrow6.8$ & $0.23\rightarrow0.31$ \\
RAFT & $960\times432$ & $1.857\rightarrow1.933$ & $+4.1\%$ & $0.604\rightarrow0.510$ & $14.5\rightarrow16.9$ & $0.52\rightarrow0.53$ \\
VideoFlow MOF & $960\times432$ & $1.727\rightarrow1.089$ & $-36.9\%$ & $0.573\rightarrow0.069$ & $4.9\rightarrow17.8$ & $0.16\rightarrow0.56$ \\
\bottomrule
\end{tabular}%
}
\end{table}

%% file: checklist.tex
\section*{NeurIPS Paper Checklist}

\begin{enumerate}

\item {\bf Claims}
    \item[] Question: Do the main claims made in the abstract and introduction accurately reflect the paper's contributions and scope?
    \item[] Answer: \answerYes{}
    \item[] Justification: The abstract and introduction state five concrete claims: (i) the FlyingDrones benchmark with dense flow and segmentation ground truth at three resolutions, (ii) the resolution-normalised EPE$_\text{n}$ metric and drone-region variants, (iii) a comparison of 16 model families, and (iv) findings on resolution-dependent separation of model families, checkpoint and curriculum dominance, and KITTI- vs.\ Sintel-style fine-tuning, plus (v) real-footage transfer and fine-tuning results, each of which is supported by the corresponding dataset section, metrics section, the result tables (Tables~\ref{tab:leaderboard-main}, \ref{tab:summary-best-drone-270x480}--\ref{tab:summary-best-drone-1080x1920} and \ref{tab:agg-full}--\ref{tab:agg-drone}), and the real-footage and fine-tuning appendices.

\item {\bf Limitations}
    \item[] Question: Does the paper discuss the limitations of the work performed by the authors?
    \item[] Answer: \answerYes{}
    \item[] Justification: The paper includes a dedicated Limitations section that discusses (i) the synthetic-only nature of the benchmark and the real-world effects it cannot capture (atmospheric turbulence, sensor noise, compression artefacts, rolling-shutter, platform-specific optics), (ii) the isolation of optical-flow quality from downstream detection or tracking performance, and (iii) the use of publicly available pretrained checkpoints with heterogeneous curricula, which means our comparison reveals the joint effect of training domain and architecture rather than a fully controlled ablation. Additional structural limitations, in particular the absence of a public pre-training dataset that jointly provides aerial ego-motion, atmospheric effects, and a small-displacement tail at $1920\times 1080$, are discussed in the Discussion and Appendix~\ref{sec:appendix-datasets}.

\item {\bf Theory assumptions and proofs}
    \item[] Question: For each theoretical result, does the paper provide the full set of assumptions and a complete (and correct) proof?
    \item[] Answer: \answerNA{}
    \item[] Justification: The paper does not present formal theorems or proofs. The methodological content is the dataset construction (Section~\ref{sec:dataset}), the precise definition of the evaluation metrics (Section~\ref{sec:methodology}: EPE, Px$\tau$, Fl, EPE$_\text{n}$, and their drone-region variants), and an empirical evaluation on pretrained checkpoints; no result depends on a theoretical claim that would require a proof.

    \item {\bf Experimental result reproducibility}
    \item[] Question: Does the paper fully disclose all the information needed to reproduce the main experimental results of the paper to the extent that it affects the main claims and/or conclusions of the paper (regardless of whether the code and data are provided or not)?
    \item[] Answer: \answerYes{}
    \item[] Justification: The dataset rendering pipeline (Blender/Cycles, multi-layer OpenEXR with RGB, Vector, and Object-Index passes), the three resolution subsets (480$\times$270, 960$\times$432, 1920$\times$1080) with summary statistics in Table~\ref{tab:dataset_stats}, the precise mathematical definition of every metric, and the exact public checkpoint names of all 16 evaluated families (with their training schedules in Table~\ref{tab:checkpoint-schedules} and source-repository URLs in Table~\ref{tab:asset_licenses}) are all described in the main paper. Each evaluated checkpoint is identified by name in the appendix tables, so any reader can re-fetch the original release and re-run the deterministic evaluation we report. The rendered FlyingDrones dataset is also published on Kaggle under \url{https://www.kaggle.com/datasets/flyingdrones/flyingdrones/data} and the code published on Github under \url{https://github.com/Znerual/flyingdrones}.

\item {\bf Open access to data and code}
    \item[] Question: Does the paper provide open access to the data and code, with sufficient instructions to faithfully reproduce the main experimental results, as described in supplemental material?
    \item[] Answer: \answerYes{}
    \item[] Justification: The FlyingDrones dataset is publicly released on Kaggle at \url{https://www.kaggle.com/datasets/flyingdrones/flyingdrones/data} with documentation and stated release limitations. The model code we evaluate is third-party and remains available under the licenses listed in Table~\ref{tab:asset_licenses}; we evaluate published checkpoints and do not redistribute third-party weights. We release the dataset loader, metric implementation, evaluation and fine-tuning splits, and per-model evaluation wrappers in a public code repository on Github under \url{https://github.com/Znerual/flyingdrones}. Per-checkpoint result tables are reported in Appendix~\ref{sec:appendix-model-results}.

\item {\bf Experimental setting/details}
    \item[] Question: Does the paper specify all the training and test details (e.g., data splits, hyperparameters, how they were chosen, type of optimizer) necessary to understand the results?
    \item[] Answer: \answerYes{}
    \item[] Justification: The benchmark evaluates published pretrained checkpoints, so the main ``setting'' is evaluation rather than training. The paper reports the evaluation subset (999 flow pairs of FlyingDrones stratified by drone size and drone motion, evaluated at each of the three resolutions, Section~\ref{sec:results-protocol-anchor}), all metrics in closed form, the exact checkpoint names and training schedules of all 16 model families (Tables~\ref{tab:checkpoint-schedules} and \ref{tab:asset_licenses}), and the inference-time choices that affect numerical results (tiled inference at 1920$\times$1080 for eight families as a fixed per-family policy, marked $\ddagger$ in the result tables). No further hyperparameters were tuned on our side: each checkpoint is run at its release default settings. The fine-tuning experiment in the appendix reports its data split, initialisation, optimiser settings, step budget, and model-selection rule.

\item {\bf Experiment statistical significance}
    \item[] Question: Does the paper report error bars suitably and correctly defined or other appropriate information about the statistical significance of the experiments?
    \item[] Answer: \answerYes{}
    \item[] Justification: Every checkpoint is evaluated with deterministic inference on a fixed stratified subset of 999 flow pairs at each resolution, and per-pair errors are stored. The leaderboard tables report 95\% percentile-bootstrap confidence intervals of the drone-region EPE (10{,}000 resamples over pairs), and paired-bootstrap probabilities for the first rank are given in the main text and appendix. Sensitivity to the choice of evaluation sample is quantified by comparing the stratified subset with a contiguous 500-frame window. Real-footage results are reported as mean $\pm$ sample standard deviation across 21 cameras, and synthetic--real agreement as rank correlations with $p$-values. Across-checkpoint variation is visible because every checkpoint of every family is reported.

\item {\bf Experiments compute resources}
    \item[] Question: For each experiment, does the paper provide sufficient information on the computer resources (type of compute workers, memory, time of execution) needed to reproduce the experiments?
    \item[] Answer: \answerYes{}
    \item[] Justification: Evaluation ran on a heterogeneous GPU cluster (NVIDIA RTX 3090, RTX 4090, RTX 5090 and RTX PRO Blackwell cards) plus a local NVIDIA RTX PRO 2000; the 264 synthetic checkpoint$\times$resolution cells took about 40--50 GPU-hours, the classical PyFlow baseline runs on CPU (several hundred CPU-hours, dominated by its $1920\times1080$ presets), the real-footage study about 10 GPU-hours, and the fine-tuning study about 5 GPU-hours (four arms of 1500 steps on a single 24--32\,GB GPU each). Eight model families are evaluated with tiled inference at $1920\times1080$ as a fixed per-family policy described in the Results section (marked $\ddagger$ in the tables). Dataset rendering used a separate headless Blender/Cycles pipeline running in parallel on GPU servers.
    
\item {\bf Code of ethics}
    \item[] Question: Does the research conducted in the paper conform, in every respect, with the NeurIPS Code of Ethics \url{https://neurips.cc/public/EthicsGuidelines}?
    \item[] Answer: \answerYes{}
    \item[] Justification: The work is based on synthetic data and benchmark analysis, and the paper explicitly discusses intended use, misuse risk, and deployment cautions in the Broader Impact section.

\item {\bf Broader impacts}
    \item[] Question: Does the paper discuss both potential positive societal impacts and negative societal impacts of the work performed?
    \item[] Answer: \answerYes{}
    \item[] Justification: The Broader Impact section discusses positive applications (airspace safety, wildlife-aware collision avoidance) and explicitly acknowledges the dual-use risk that drone-analysis data can be used in surveillance settings. It recommends application-specific validation, human oversight, and clearly documented operational limits in any downstream deployment.
    
\item {\bf Safeguards}
    \item[] Question: Does the paper describe safeguards that have been put in place for responsible release of data or models that have a high risk for misuse (e.g., pre-trained language models, image generators, or scraped datasets)?
    \item[] Answer: \answerNA{}
    \item[] Justification: The submission does not release a model. It introduces a synthetic dataset for evaluating optical-flow estimation on small aerial targets; the imagery is fully simulated and contains no personal or sensitive content. We do not consider the resulting misuse risk to be unusually high relative to standard computer-vision benchmark releases, so additional release safeguards are not required beyond the documentation, intended-use guidance, and operational-limit recommendations stated in the Broader Impact section.

\item {\bf Licenses for existing assets}
    \item[] Question: Are the creators or original owners of assets (e.g., code, data, models), used in the paper, properly credited and are the license and terms of use explicitly mentioned and properly respected?
    \item[] Answer: \answerYes{}
    \item[] Justification: Table~\ref{tab:asset_licenses} (Appendix~\ref{sec:appendix-licenses}) lists the official source URL and the license or public terms for every external asset used in the study: datasets (FlyingChairs, FlyingThings3D, MPI Sintel, KITTI, TartanAir, Spring, HD1K) and model repositories (FlowNet2, PyFlow, PWC-Net, LiteFlowNet3, VCN, ARFlow, FastFlowNet, RAFT, GMFlow, UniMatch, FlowFormer++, VideoFlow, FlowDiffusion/DDVM, MemFlow, SEA-RAFT, WAFT, plus the DepthAnythingV2 encoder used by WAFT). The table explicitly states cases where no standalone public license is exposed (e.g.\ HD1K) and cases where the code license is more permissive than the dataset terms (e.g.\ FlyingChairs/FlyingThings3D research-only use). We evaluate published checkpoints and do not redistribute third-party datasets or weights as part of this submission.

\item {\bf New assets}
    \item[] Question: Are new assets introduced in the paper well documented and is the documentation provided alongside the assets?
    \item[] Answer: \answerYes{}
    \item[] Justification: The submission introduces FlyingDrones, a new synthetic optical-flow benchmark, publicly hosted on Kaggle at \url{https://www.kaggle.com/datasets/flyingdrones/flyingdrones/data}. The main paper documents the rendering pipeline (Blender/Cycles, headless GPU rendering, multi-layer OpenEXR export of RGB, raw Vector pass, and Object Index pass), the post-processing that yields per-pixel motion vectors and per-drone instance masks, the three resolution subsets, the dataset summary statistics (Table~\ref{tab:dataset_stats}; 9{,}001 frames, 19{,}999 drone instances, median target area $155$\,px), and the metrics defined on top of it. The Kaggle release ships with documentation and release limitations.

\item {\bf Crowdsourcing and research with human subjects}
    \item[] Question: For crowdsourcing experiments and research with human subjects, does the paper include the full text of instructions given to participants and screenshots, if applicable, as well as details about compensation (if any)? 
    \item[] Answer: \answerNA{}
    \item[] Justification: The work does not involve crowdsourcing or human subjects.

\item {\bf Institutional review board (IRB) approvals or equivalent for research with human subjects}
    \item[] Question: Does the paper describe potential risks incurred by study participants, whether such risks were disclosed to the subjects, and whether Institutional Review Board (IRB) approvals (or an equivalent approval/review based on the requirements of your country or institution) were obtained?
    \item[] Answer: \answerNA{}
    \item[] Justification: The work does not involve crowdsourcing or human subjects, so IRB-style review is not applicable.

\item {\bf Declaration of LLM usage}
    \item[] Question: Does the paper describe the usage of LLMs if it is an important, original, or non-standard component of the core methods in this research? Note that if the LLM is used only for writing, editing, or formatting purposes and does \emph{not} impact the core methodology, scientific rigor, or originality of the research, declaration is not required.
    \item[] Answer: \answerNA{}
    \item[] Justification: LLMs are not part of the research methodology. The contributions (dataset construction, metrics, and the benchmark of 16 optical-flow model families) do not use LLMs as a component.

\end{enumerate}

%% file: references.bib
@inproceedings{kondermann2016,
	title = {The {HCI} Benchmark Suite: Stereo and Flow Ground Truth With Uncertainties for Urban Autonomous Driving},
	pages = {19--28},
	booktitle = {Proceedings of the {IEEE} Conference on Computer Vision and Pattern Recognition Workshops},
	author = {Kondermann, Daniel and Nair, Rahul and Honauer, Katrin and Krispin, Karsten and Andrulis, Jonas and Brock, Alexander and Gussefeld, Burkhard and Rahimimoghaddam, Mohsen and Hofmann, Sabine and Brenner, Claus and {others}},
	date = {2016},
}

@inproceedings{cordts2016,
	title = {The Cityscapes Dataset for Semantic Urban Scene Understanding},
	booktitle = {Proc. of the {IEEE} Conference on Computer Vision and Pattern Recognition ({CVPR})},
	author = {Cordts, Marius and Omran, Mohamed and Ramos, Sebastian and Rehfeld, Timo and Enzweiler, Markus and Benenson, Rodrigo and Franke, Uwe and Roth, Stefan and Schiele, Bernt},
	date = {2016},
}

@inproceedings{mehl2023,
	title = {Spring: A High-Resolution High-Detail Dataset and Benchmark for Scene Flow, Optical Flow and Stereo},
	pages = {4981--4991},
	booktitle = {Proc. {IEEE}/{CVF} Conference on Computer Vision and Pattern Recognition ({CVPR})},
	author = {Mehl, Lukas and Schmalfuss, Jenny and Jahedi, Azin and Nalivayko, Yaroslava and Bruhn, Andrés},
	date = {2023},
}

@article{mahalanobis2025,
	title = {Efficient Drone Detection Using Temporal Anomalies and Small Spatio-Temporal Networks},
	volume = {26},
	issn = {1424-8220},
	url = {https://pmc.ncbi.nlm.nih.gov/articles/PMC12788262/},
	doi = {10.3390/s26010170},
	pages = {170},
	number = {1},
	journaltitle = {Sensors (Basel, Switzerland)},
	shortjournal = {Sensors (Basel)},
	author = {Mahalanobis, Abhijit and Tall, Amadou},
	urldate = {2026-05-04},
	date = {2025-12-26},
}

@inproceedings{srigrarom2020,
	title = {Hybrid motion-based object detection for detecting and tracking of small and fast moving drones},
	issn = {2575-7296},
	url = {https://ieeexplore.ieee.org/document/9213912},
	doi = {10.1109/ICUAS48674.2020.9213912},
	eventtitle = {2020 International Conference on Unmanned Aircraft Systems ({ICUAS})},
	pages = {615--621},
	booktitle = {2020 International Conference on Unmanned Aircraft Systems ({ICUAS})},
	author = {Srigrarom, Sutthiphong and Hoe Chew, Kim},
	urldate = {2026-05-04},
	date = {2020-09},
	note = {{ISSN}: 2575-7296},
}

@misc{li_reconstruction_2020,
	title = {Reconstruction of 3D flight trajectories from ad-hoc camera networks},
	url = {http://arxiv.org/abs/2003.04784},
	doi = {10.48550/arXiv.2003.04784},
	number = {{arXiv}:2003.04784},
	publisher = {{arXiv}},
	author = {Li, Jingtong and Murray, Jesse and Ismaili, Dorina and Schindler, Konrad and Albl, Cenek},
	urldate = {2026-07-27},
	date = {2020-07-29},
	eprinttype = {arxiv},
	eprint = {2003.04784 [cs.RO]},
}

@inproceedings{mayer_large_2016,
	title = {A Large Dataset to Train Convolutional Networks for Disparity, Optical Flow, and Scene Flow Estimation},
	url = {http://lmb.informatik.uni-freiburg.de/Publications/2016/MIFDB16},
	booktitle = {{IEEE} International Conference on Computer Vision and Pattern Recognition ({CVPR})},
	author = {Mayer, N. and Ilg, E. and Häusser, P. and Fischer, P. and Cremers, D. and Dosovitskiy, A. and Brox, T.},
	date = {2016},
}

@article{cheng_towards_2023,
	title = {Towards Large-Scale Small Object Detection: Survey and Benchmarks},
	volume = {45},
	issn = {1939-3539},
	url = {https://ieeexplore.ieee.org/document/10168277/},
	doi = {10.1109/TPAMI.2023.3290594},
	shorttitle = {Towards Large-Scale Small Object Detection},
	pages = {13467--13488},
	number = {11},
	journaltitle = {{IEEE} Transactions on Pattern Analysis and Machine Intelligence},
	author = {Cheng, Gong and Yuan, Xiang and Yao, Xiwen and Yan, Kebing and Zeng, Qinghua and Xie, Xingxing and Han, Junwei},
	urldate = {2026-04-30},
	date = {2023-11},
}

@article{liu_survey_2021,
	title = {A survey and performance evaluation of deep learning methods for small object detection},
	volume = {172},
	issn = {0957-4174},
	url = {https://www.sciencedirect.com/science/article/pii/S0957417421000439},
	doi = {10.1016/j.eswa.2021.114602},
	pages = {114602},
	journaltitle = {Expert Systems with Applications},
	shortjournal = {Expert Systems with Applications},
	author = {Liu, Yang and Sun, Peng and Wergeles, Nickolas and Shang, Yi},
	urldate = {2026-04-30},
	date = {2021-06-15},
}

@article{tong_recent_2020,
	title = {Recent advances in small object detection based on deep learning: A review},
	volume = {97},
	issn = {0262-8856},
	url = {https://www.sciencedirect.com/science/article/pii/S0262885620300421},
	doi = {10.1016/j.imavis.2020.103910},
	shorttitle = {Recent advances in small object detection based on deep learning},
	pages = {103910},
	journaltitle = {Image and Vision Computing},
	shortjournal = {Image and Vision Computing},
	author = {Tong, Kang and Wu, Yiquan and Zhou, Fei},
	urldate = {2026-04-30},
	date = {2020-05-01},
}

@misc{lin_feature_2017,
	title = {Feature Pyramid Networks for Object Detection},
	url = {http://arxiv.org/abs/1612.03144},
	doi = {10.48550/arXiv.1612.03144},
	number = {{arXiv}:1612.03144},
	publisher = {{arXiv}},
	author = {Lin, Tsung-Yi and Dollár, Piotr and Girshick, Ross and He, Kaiming and Hariharan, Bharath and Belongie, Serge},
	urldate = {2026-04-30},
	date = {2017-04-19},
	eprinttype = {arxiv},
	eprint = {1612.03144 [cs]},
}

@article{sun_enhancing_2023,
	title = {Enhancing {UAV} Detection in Surveillance Camera Videos through Spatiotemporal Information and Optical Flow},
	volume = {23},
	rights = {http://creativecommons.org/licenses/by/3.0/},
	issn = {1424-8220},
	url = {https://www.mdpi.com/1424-8220/23/13/6037},
	doi = {10.3390/s23136037},
	pages = {6037},
	number = {13},
	journaltitle = {Sensors},
	publisher = {Multidisciplinary Digital Publishing Institute},
	author = {Sun, Yu and Zhi, Xiyang and Han, Haowen and Jiang, Shikai and Shi, Tianjun and Gong, Jinnan and Zhang, Wei},
	urldate = {2026-04-30},
	date = {2023-01},
	langid = {english},
}

@article{alkentar_utilize_2024,
	title = {Utilize Dense Optical Flow for Small Flying Targets Detection and Tracking},
	volume = {98},
	rights = {Copyright (c) 2024 American Scientific Research Journal for Engineering, Technology, and Sciences},
	issn = {2313-4402},
	url = {https://asrjetsjournal.org/American_Scientific_Journal/article/view/10281},
	pages = {53--70},
	number = {1},
	journaltitle = {American Scientific Research Journal for Engineering, Technology, and Sciences},
	author = {Alkentar, Saad and Assalem, Abdulkareem},
	urldate = {2026-04-30},
	date = {2024-05-26},
	langid = {english},
}

@inproceedings{saxena_surprising_2023,
	title = {The Surprising Effectiveness of Diffusion Models for Optical Flow and Monocular Depth Estimation},
	url = {https://openreview.net/forum?id=jDIlzSU8wJ},
	booktitle = {Thirty-seventh Conference on Neural Information Processing Systems},
	author = {Saxena, Saurabh and Herrmann, Charles and Hur, Junhwa and Kar, Abhishek and Norouzi, Mohammad and Sun, Deqing and Fleet, David J.},
	date = {2023},
}

@inproceedings{pathak_learning_2017,
	title = {Learning Features by Watching Objects Move},
	booktitle = {Computer Vision and Pattern Recognition ({CVPR})},
	author = {Pathak, Deepak and Girshick, Ross and Dollár, Piotr and Darrell, Trevor and Hariharan, Bharath},
	date = {2017},
}

@article{shi_videoflow_2023,
	title = {Videoflow: Exploiting temporal cues for multi-frame optical flow estimation},
	journaltitle = {{arXiv} preprint {arXiv}:2303.08340},
	author = {Shi, Xiaoyu and Huang, Zhaoyang and Bian, Weikang and Li, Dasong and Zhang, Manyuan and Cheung, Ka Chun and See, Simon and Qin, Hongwei and Dai, Jifeng and Li, Hongsheng},
	date = {2023},
}

@inproceedings{yang_volumetric_2019,
	title = {Volumetric Correspondence Networks for Optical Flow},
	booktitle = {{NeurIPS}},
	author = {Yang, Gengshan and Ramanan, Deva},
	date = {2019},
}

@article{xu_unifying_2023,
	title = {Unifying Flow, Stereo and Depth Estimation},
	journaltitle = {{IEEE} Transactions on Pattern Analysis and Machine Intelligence},
	author = {Xu, Haofei and Zhang, Jing and Cai, Jianfei and Rezatofighi, Hamid and Yu, Fisher and Tao, Dacheng and Geiger, Andreas},
	date = {2023},
}

@inproceedings{dong_memflow_2024,
	title = {{MemFlow}: Optical Flow Estimation and Prediction with Memory},
	booktitle = {Proceedings of the {IEEE}/{CVF} Conference on Computer Vision and Pattern Recognition},
	author = {Dong, Qiaole and Fu, Yanwei},
	date = {2024},
}

@inproceedings{hui_liteflownet3_2020,
	title = {{LiteFlowNet}3: Resolving Correspondence Ambiguity for More Accurate Optical Flow Estimation},
	pages = {169--184},
	booktitle = {Proceedings of the European Conference on Computer Vision ({ECCV})},
	author = {Hui, Tak-Wai and Loy, Chen Change},
	date = {2020},
}

@inproceedings{xu_gmflow_2022,
	title = {{GMFlow}: Learning Optical Flow via Global Matching},
	pages = {8121--8130},
	booktitle = {Proceedings of the {IEEE}/{CVF} Conference on Computer Vision and Pattern Recognition},
	author = {Xu, Haofei and Zhang, Jing and Cai, Jianfei and Rezatofighi, Hamid and Tao, Dacheng},
	date = {2022},
}

@inproceedings{shi_flowformer_2023,
	title = {Flowformer++: Masked cost volume autoencoding for pretraining optical flow estimation},
	pages = {1599--1610},
	booktitle = {Proceedings of the {IEEE}/{CVF} Conference on Computer Vision and Pattern Recognition},
	author = {Shi, Xiaoyu and Huang, Zhaoyang and Li, Dasong and Zhang, Manyuan and Cheung, Ka Chun and See, Simon and Qin, Hongwei and Dai, Jifeng and Li, Hongsheng},
	date = {2023},
}

@misc{dong_open-ddvm_2023,
	title = {Open-{DDVM}: A Reproduction and Extension of Diffusion Model for Optical Flow Estimation},
	author = {Dong, Qiaole and Zhao, Bo and Fu, Yanwei},
	date = {2023},
	note = {Publication Title: {arXiv} preprint {arXiv}:2312.01746},
}

@inproceedings{kong_fastflownet_2021,
	title = {{FastFlowNet}: A Lightweight Network for Fast Optical Flow Estimation},
	booktitle = {2021 {IEEE} International Conference on Robotics and Automation ({ICRA})},
	author = {Kong, Lingtong and Shen, Chunhua and Yang, Jie},
	date = {2021},
}

@inproceedings{liu_learning_2020,
	title = {Learning by Analogy: Reliable Supervision from Transformations for Unsupervised Optical Flow Estimation},
	booktitle = {{IEEE} Conference on Computer Vision and Pattern Recognition({CVPR})},
	author = {Liu, Liang and Zhang, Jiangning and He, Ruifei and Liu, Yong and Wang, Yabiao and Tai, Ying and Luo, Donghao and Wang, Chengjie and Li, Jilin and Huang, Feiyue},
	date = {2020},
}

@article{tartanair2020iros,
	title = {{TartanAir}: a dataset to push the limits of visual {SLAM}},
	journaltitle = {2020 {IEEE}/{RSJ} international conference on intelligent robots and systems ({IROS})},
	author = {Wang, Wenshan and Zhu, Delong and Wang, Xiangwei and Hu, Yaoyu and Qiu, Yuheng and Wang, Chen and Hu, Yafei and Kapoor, Ashish and Scherer, Sebastian},
	date = {2020},
}

@inproceedings{Menze2015CVPR,
	title = {Object scene flow for autonomous vehicles},
	booktitle = {Conference on computer vision and pattern recognition ({CVPR})},
	author = {Menze, Moritz and Geiger, Andreas},
	date = {2015},
}

@inproceedings{Butler:ECCV:2012,
	title = {A naturalistic open source movie for optical flow evaluation},
	series = {Part {IV}, {LNCS} 7577},
	pages = {611--625},
	booktitle = {European conf. on computer vision ({ECCV})},
	publisher = {Springer-Verlag},
	author = {Butler, D. J. and Wulff, J. and Stanley, G. B. and Black, M. J.},
	editor = {{A. Fitzgibbon et al. (Eds.)}},
	date = {2012-10},
}

@article{depth_anything_v2,
	title = {Depth anything V2},
	journaltitle = {{arXiv}:2406.09414},
	author = {Yang, Lihe and Kang, Bingyi and Huang, Zilong and Zhao, Zhen and Xu, Xiaogang and Feng, Jiashi and Zhao, Hengshuang},
	date = {2024},
}

@article{ren2017,
	title = {Faster R-{CNN}: Towards Real-Time Object Detection with Region Proposal Networks},
	volume = {39},
	rights = {https://ieeexplore.ieee.org/Xplorehelp/downloads/license-information/{IEEE}.html},
	issn = {0162-8828, 2160-9292, 1939-3539},
	url = {https://ieeexplore.ieee.org/document/7485869/},
	doi = {10.1109/TPAMI.2016.2577031},
	shorttitle = {Faster R-{CNN}},
	pages = {1137--1149},
	number = {6},
	journaltitle = {{IEEE} Transactions on Pattern Analysis and Machine Intelligence},
	shortjournal = {{IEEE} Trans. Pattern Anal. Mach. Intell.},
	author = {Ren, Shaoqing and He, Kaiming and Girshick, Ross and Sun, Jian},
	urldate = {2026-01-28},
	date = {2017-06-01},
	langid = {english},
}

@inproceedings{redmon2016a,
	location = {Las Vegas, {NV}, {USA}},
	title = {You Only Look Once: Unified, Real-Time Object Detection},
	isbn = {978-1-4673-8851-1},
	url = {http://ieeexplore.ieee.org/document/7780460/},
	doi = {10.1109/CVPR.2016.91},
	shorttitle = {You Only Look Once},
	eventtitle = {2016 {IEEE} Conference on Computer Vision and Pattern Recognition ({CVPR})},
	pages = {779--788},
	booktitle = {2016 {IEEE} Conference on Computer Vision and Pattern Recognition ({CVPR})},
	publisher = {{IEEE}},
	author = {Redmon, Joseph and Divvala, Santosh and Girshick, Ross and Farhadi, Ali},
	urldate = {2026-01-28},
	date = {2016-06},
	langid = {english},
}

@misc{wang2025,
	title = {{WAFT}: Warping-Alone Field Transforms for Optical Flow},
	url = {http://arxiv.org/abs/2506.21526},
	doi = {10.48550/arXiv.2506.21526},
	shorttitle = {{WAFT}},
	number = {{arXiv}:2506.21526},
	publisher = {{arXiv}},
	author = {Wang, Yihan and Deng, Jia},
	urldate = {2026-01-28},
	date = {2025-10-07},
	eprinttype = {arxiv},
	eprint = {2506.21526 [cs]},
}

@inproceedings{wang2025a,
	location = {Cham},
	title = {{SEA}-{RAFT}: Simple, Efficient, Accurate {RAFT} for Optical Flow},
	volume = {15065},
	isbn = {978-3-031-72666-8 978-3-031-72667-5},
	url = {https://link.springer.com/10.1007/978-3-031-72667-5_3},
	doi = {10.1007/978-3-031-72667-5_3},
	shorttitle = {{SEA}-{RAFT}},
	pages = {36--54},
	booktitle = {Computer Vision – {ECCV} 2024},
	publisher = {Springer Nature Switzerland},
	author = {Wang, Yihan and Lipson, Lahav and Deng, Jia},
	editor = {Leonardis, Aleš and Ricci, Elisa and Roth, Stefan and Russakovsky, Olga and Sattler, Torsten and Varol, Gül},
	urldate = {2026-01-28},
	date = {2025},
	langid = {english},
	note = {Series Title: Lecture Notes in Computer Science},
}

@inproceedings{teed2020,
	location = {Cham},
	title = {{RAFT}: Recurrent All-Pairs Field Transforms for Optical Flow},
	volume = {12347},
	isbn = {978-3-030-58535-8 978-3-030-58536-5},
	url = {https://link.springer.com/10.1007/978-3-030-58536-5_24},
	doi = {10.1007/978-3-030-58536-5_24},
	shorttitle = {{RAFT}},
	pages = {402--419},
	booktitle = {Computer Vision – {ECCV} 2020},
	publisher = {Springer International Publishing},
	author = {Teed, Zachary and Deng, Jia},
	editor = {Vedaldi, Andrea and Bischof, Horst and Brox, Thomas and Frahm, Jan-Michael},
	urldate = {2026-01-28},
	date = {2020},
	langid = {english},
	note = {Series Title: Lecture Notes in Computer Science},
}

@inproceedings{ilg2017,
	location = {Honolulu, {HI}},
	title = {{FlowNet} 2.0: Evolution of Optical Flow Estimation with Deep Networks},
	isbn = {978-1-5386-0457-1},
	url = {http://ieeexplore.ieee.org/document/8099662/},
	doi = {10.1109/CVPR.2017.179},
	shorttitle = {{FlowNet} 2.0},
	eventtitle = {2017 {IEEE} Conference on Computer Vision and Pattern Recognition ({CVPR})},
	pages = {1647--1655},
	booktitle = {2017 {IEEE} Conference on Computer Vision and Pattern Recognition ({CVPR})},
	publisher = {{IEEE}},
	author = {Ilg, Eddy and Mayer, Nikolaus and Saikia, Tonmoy and Keuper, Margret and Dosovitskiy, Alexey and Brox, Thomas},
	urldate = {2026-01-28},
	date = {2017-07},
	langid = {english},
}

@inproceedings{dosovitskiy2015,
	location = {Santiago},
	title = {{FlowNet}: Learning Optical Flow with Convolutional Networks},
	isbn = {978-1-4673-8391-2},
	url = {https://ieeexplore.ieee.org/document/7410673/},
	doi = {10.1109/ICCV.2015.316},
	shorttitle = {{FlowNet}},
	eventtitle = {2015 {IEEE} International Conference on Computer Vision ({ICCV})},
	pages = {2758--2766},
	booktitle = {2015 {IEEE} International Conference on Computer Vision ({ICCV})},
	publisher = {{IEEE}},
	author = {Dosovitskiy, Alexey and Fischer, Philipp and Ilg, Eddy and Hausser, Philip and Hazirbas, Caner and Golkov, Vladimir and Smagt, Patrick Van Der and Cremers, Daniel and Brox, Thomas},
	urldate = {2026-01-28},
	date = {2015-12},
	langid = {english},
}

@inproceedings{sun2018,
	location = {Salt Lake City, {UT}, {USA}},
	title = {{PWC}-Net: {CNNs} for Optical Flow Using Pyramid, Warping, and Cost Volume},
	isbn = {978-1-5386-6420-9},
	url = {https://ieeexplore.ieee.org/document/8579029/},
	doi = {10.1109/CVPR.2018.00931},
	shorttitle = {{PWC}-Net},
	eventtitle = {2018 {IEEE}/{CVF} Conference on Computer Vision and Pattern Recognition ({CVPR})},
	pages = {8934--8943},
	booktitle = {2018 {IEEE}/{CVF} Conference on Computer Vision and Pattern Recognition},
	publisher = {{IEEE}},
	author = {Sun, Deqing and Yang, Xiaodong and Liu, Ming-Yu and Kautz, Jan},
	urldate = {2026-01-28},
	date = {2018-06},
	langid = {english},
}

@inproceedings{hui18liteflownet,
	title = {{LiteFlowNet}: a lightweight convolutional neural network for optical flow estimation},
	pages = {8981--8989},
	booktitle = {Proceedings of {IEEE} conference on computer vision and pattern recognition ({CVPR})},
	author = {Hui, Tak-Wai and Tang, Xiaoou and Loy, Chen Change},
	date = {2018-06},
}

@inproceedings{lucas1981,
	title = {An Iterative Image Registration Technique with an Application to Stereo Vision ({IJCAI})},
	volume = {81},
	eventtitle = {[No source information available]},
	author = {Lucas, Bruce and Kanade, Takeo},
	date = {1981-04-01},
}

@article{horn1981,
	title = {Determining optical flow},
	volume = {17},
	issn = {0004-3702},
	url = {https://www.sciencedirect.com/science/article/pii/0004370281900242},
	doi = {10.1016/0004-3702(81)90024-2},
	pages = {185--203},
	number = {1},
	journaltitle = {Artificial Intelligence},
	shortjournal = {Artificial Intelligence},
	author = {Horn, Berthold K. P. and Schunck, Brian G.},
	urldate = {2024-09-27},
	date = {1981-08-01},
}
